%!PS-Adobe-3.0
/Win35Dict 290 dict def Win35Dict begin/bd{bind def}bind def/in{72
mul}bd/ed{exch def}bd/ld{load def}bd/tr/translate ld/gs/gsave ld/gr
/grestore ld/M/moveto ld/L/lineto ld/rmt/rmoveto ld/rlt/rlineto ld
/rct/rcurveto ld/st/stroke ld/n/newpath ld/sm/setmatrix ld/cm/currentmatrix
ld/cp/closepath ld/ARC/arcn ld/TR{65536 div}bd/lj/setlinejoin ld/lc
/setlinecap ld/ml/setmiterlimit ld/sl/setlinewidth ld/scignore false
def/sc{scignore{pop pop pop}{0 index 2 index eq 2 index 4 index eq
and{pop pop 255 div setgray}{3{255 div 3 1 roll}repeat setrgbcolor}ifelse}ifelse}bd
/FC{bR bG bB sc}bd/fC{/bB ed/bG ed/bR ed}bd/HC{hR hG hB sc}bd/hC{
/hB ed/hG ed/hR ed}bd/PC{pR pG pB sc}bd/pC{/pB ed/pG ed/pR ed}bd/sM
matrix def/PenW 1 def/iPen 5 def/mxF matrix def/mxE matrix def/mxUE
matrix def/mxUF matrix def/fBE false def/iDevRes 72 0 matrix defaultmatrix
dtransform dup mul exch dup mul add sqrt def/fPP false def/SS{fPP{
/SV save def}{gs}ifelse}bd/RS{fPP{SV restore}{gr}ifelse}bd/EJ{gsave
showpage grestore}bd/#C{userdict begin/#copies ed end}bd/FEbuf 2 string
def/FEglyph(G  )def/FE{1 exch{dup 16 FEbuf cvrs FEglyph exch 1 exch
putinterval 1 index exch FEglyph cvn put}for}bd/SM{/iRes ed/cyP ed
/cxPg ed/cyM ed/cxM ed 72 100 div dup scale dup 0 ne{90 eq{cyM exch
0 eq{cxM exch tr -90 rotate -1 1 scale}{cxM cxPg add exch tr +90 rotate}ifelse}{cyP
cyM sub exch 0 ne{cxM exch tr -90 rotate}{cxM cxPg add exch tr -90
rotate 1 -1 scale}ifelse}ifelse}{pop cyP cyM sub exch 0 ne{cxM cxPg
add exch tr 180 rotate}{cxM exch tr 1 -1 scale}ifelse}ifelse 100 iRes
div dup scale 0 0 transform .25 add round .25 sub exch .25 add round
.25 sub exch itransform translate}bd/SJ{1 index 0 eq{pop pop/fBE false
def}{1 index/Break ed div/dxBreak ed/fBE true def}ifelse}bd/ANSIVec[
16#0/grave 16#1/acute 16#2/circumflex 16#3/tilde 16#4/macron 16#5/breve
16#6/dotaccent 16#7/dieresis 16#8/ring 16#9/cedilla 16#A/hungarumlaut
16#B/ogonek 16#C/caron 16#D/dotlessi 16#27/quotesingle 16#60/grave
16#7C/bar 16#82/quotesinglbase 16#83/florin 16#84/quotedblbase 16#85
/ellipsis 16#86/dagger 16#87/daggerdbl 16#88/circumflex 16#89/perthousand
16#8A/Scaron 16#8B/guilsinglleft 16#8C/OE 16#91/quoteleft 16#92/quoteright
16#93/quotedblleft 16#94/quotedblright 16#95/bullet 16#96/endash 16#97
/emdash 16#98/tilde 16#99/trademark 16#9A/scaron 16#9B/guilsinglright
16#9C/oe 16#9F/Ydieresis 16#A0/space 16#A1/exclamdown 16#A4/currency
16#A5/yen 16#A6/brokenbar 16#A7/section 16#A8/dieresis 16#A9/copyright
16#AA/ordfeminine 16#AB/guillemotleft 16#AC/logicalnot 16#AD/hyphen
16#AE/registered 16#AF/macron 16#B0/degree 16#B1/plusminus 16#B2/twosuperior
16#B3/threesuperior 16#B4/acute 16#B5/mu 16#B6/paragraph 16#B7/periodcentered
16#B8/cedilla 16#B9/onesuperior 16#BA/ordmasculine 16#BB/guillemotright
16#BC/onequarter 16#BD/onehalf 16#BE/threequarters 16#BF/questiondown
16#C0/Agrave 16#C1/Aacute 16#C2/Acircumflex 16#C3/Atilde 16#C4/Adieresis
16#C5/Aring 16#C6/AE 16#C7/Ccedilla 16#C8/Egrave 16#C9/Eacute 16#CA
/Ecircumflex 16#CB/Edieresis 16#CC/Igrave 16#CD/Iacute 16#CE/Icircumflex
16#CF/Idieresis 16#D0/Eth 16#D1/Ntilde 16#D2/Ograve 16#D3/Oacute 16#D4
/Ocircumflex 16#D5/Otilde 16#D6/Odieresis 16#D7/multiply 16#D8/Oslash
16#D9/Ugrave 16#DA/Uacute 16#DB/Ucircumflex 16#DC/Udieresis 16#DD/Yacute
16#DE/Thorn 16#DF/germandbls 16#E0/agrave 16#E1/aacute 16#E2/acircumflex
16#E3/atilde 16#E4/adieresis 16#E5/aring 16#E6/ae 16#E7/ccedilla 16#E8
/egrave 16#E9/eacute 16#EA/ecircumflex 16#EB/edieresis 16#EC/igrave
16#ED/iacute 16#EE/icircumflex 16#EF/idieresis 16#F0/eth 16#F1/ntilde
16#F2/ograve 16#F3/oacute 16#F4/ocircumflex 16#F5/otilde 16#F6/odieresis
16#F7/divide 16#F8/oslash 16#F9/ugrave 16#FA/uacute 16#FB/ucircumflex
16#FC/udieresis 16#FD/yacute 16#FE/thorn 16#FF/ydieresis ] def/reencdict
12 dict def/IsChar{basefontdict/CharStrings get exch known}bd/MapCh{dup
IsChar not{pop/bullet}if newfont/Encoding get 3 1 roll put}bd/MapDegree{16#b0
/degree IsChar{/degree}{/ring}ifelse MapCh}bd/MapBB{16#a6/brokenbar
IsChar{/brokenbar}{/bar}ifelse MapCh}bd/ANSIFont{reencdict begin/newfontname
ed/basefontname ed FontDirectory newfontname known not{/basefontdict
basefontname findfont def/newfont basefontdict maxlength dict def basefontdict{exch
dup/FID ne{dup/Encoding eq{exch dup length array copy newfont 3 1 roll
put}{exch newfont 3 1 roll put}ifelse}{pop pop}ifelse}forall newfont
/FontName newfontname put 127 1 159{newfont/Encoding get exch/bullet
put}for ANSIVec aload pop ANSIVec length 2 idiv{MapCh}repeat MapDegree
MapBB newfontname newfont definefont pop}if newfontname end}bd/SB{FC
/ULlen ed/str ed str length fBE not{dup 1 gt{1 sub}if}if/cbStr ed
/dxGdi ed/y0 ed/x0 ed str stringwidth dup 0 ne{/y1 ed/x1 ed y1 y1
mul x1 x1 mul add sqrt dxGdi exch div 1 sub dup x1 mul cbStr div exch
y1 mul cbStr div}{exch abs neg dxGdi add cbStr div exch}ifelse/dyExtra
ed/dxExtra ed x0 y0 M fBE{dxBreak 0 BCh dxExtra dyExtra str awidthshow}{dxExtra
dyExtra str ashow}ifelse fUL{x0 y0 M dxUL dyUL rmt ULlen fBE{Break
add}if 0 mxUE transform gs rlt cyUL sl [] 0 setdash st gr}if fSO{x0
y0 M dxSO dySO rmt ULlen fBE{Break add}if 0 mxUE transform gs rlt cyUL
sl [] 0 setdash st gr}if n/fBE false def}bd/font{/name ed/Ascent ed
0 ne/fT3 ed 0 ne/fSO ed 0 ne/fUL ed/Sy ed/Sx ed 10.0 div/ori ed -10.0
div/esc ed/BCh ed name findfont/xAscent 0 def/yAscent Ascent def/ULesc
esc def ULesc mxUE rotate pop fT3{/esc 0 def xAscent yAscent mxUE transform
/yAscent ed/xAscent ed}if [Sx 0 0 Sy neg xAscent yAscent] esc mxE
rotate mxF concatmatrix makefont setfont [Sx 0 0 Sy neg 0 Ascent] mxUE
mxUF concatmatrix pop fUL{currentfont dup/FontInfo get/UnderlinePosition
known not{pop/Courier findfont}if/FontInfo get/UnderlinePosition get
1000 div 0 exch mxUF transform/dyUL ed/dxUL ed}if fSO{0 .3 mxUF transform
/dySO ed/dxSO ed}if fUL fSO or{currentfont dup/FontInfo get/UnderlineThickness
known not{pop/Courier findfont}if/FontInfo get/UnderlineThickness get
1000 div Sy mul/cyUL ed}if}bd/min{2 copy gt{exch}if pop}bd/max{2 copy
lt{exch}if pop}bd/CP{/ft ed{{ft 0 eq{clip}{eoclip}ifelse}stopped{currentflat
1 add setflat}{exit}ifelse}loop}bd/patfont 10 dict def patfont begin
/FontType 3 def/FontMatrix [1 0 0 -1 0 0] def/FontBBox [0 0 16 16]
def/Encoding StandardEncoding def/BuildChar{pop pop 16 0 0 0 16 16
setcachedevice 16 16 false [1 0 0 1 .25 .25]{pat}imagemask}bd end/p{
/pat 32 string def{}forall 0 1 7{dup 2 mul pat exch 3 index put dup
2 mul 1 add pat exch 3 index put dup 2 mul 16 add pat exch 3 index
put 2 mul 17 add pat exch 2 index put pop}for}bd/pfill{/PatFont patfont
definefont setfont/ch(AAAA)def X0 64 X1{Y1 -16 Y0{1 index exch M ch
show}for pop}for}bd/vert{X0 w X1{dup Y0 M Y1 L st}for}bd/horz{Y0 w
Y1{dup X0 exch M X1 exch L st}for}bd/fdiag{X0 w X1{Y0 M X1 X0 sub dup
rlt st}for Y0 w Y1{X0 exch M Y1 Y0 sub dup rlt st}for}bd/bdiag{X0 w
X1{Y1 M X1 X0 sub dup neg rlt st}for Y0 w Y1{X0 exch M Y1 Y0 sub dup
neg rlt st}for}bd/AU{1 add cvi 15 or}bd/AD{1 sub cvi -16 and}bd/SHR{pathbbox
AU/Y1 ed AU/X1 ed AD/Y0 ed AD/X0 ed}bd/hfill{/w iRes 37.5 div round
def 0.1 sl [] 0 setdash n dup 0 eq{horz}if dup 1 eq{vert}if dup 2 eq{fdiag}if
dup 3 eq{bdiag}if dup 4 eq{horz vert}if 5 eq{fdiag bdiag}if}bd/F{/ft
ed fm 256 and 0 ne{gs FC ft 0 eq{fill}{eofill}ifelse gr}if fm 1536
and 0 ne{SHR gs HC ft CP fm 1024 and 0 ne{/Tmp save def pfill Tmp restore}{fm
15 and hfill}ifelse gr}if}bd/S{PenW sl PC st}bd/m matrix def/GW{iRes
12 div PenW add cvi}bd/DoW{iRes 50 div PenW add cvi}bd/DW{iRes 8 div
PenW add cvi}bd/SP{/PenW ed/iPen ed iPen 0 eq iPen 6 eq or{[] 0 setdash}if
iPen 1 eq{[DW GW] 0 setdash}if iPen 2 eq{[DoW GW] 0 setdash}if iPen
3 eq{[DW GW DoW GW] 0 setdash}if iPen 4 eq{[DW GW DoW GW DoW GW] 0
setdash}if}bd/E{m cm pop tr scale 1 0 moveto 0 0 1 0 360 arc cp m sm}bd
/AG{/sy ed/sx ed sx div 4 1 roll sy div 4 1 roll sx div 4 1 roll sy
div 4 1 roll atan/a2 ed atan/a1 ed sx sy scale a1 a2 ARC}def/A{m cm
pop tr AG m sm}def/P{m cm pop tr 0 0 M AG cp m sm}def/RRect{n 4 copy
M 3 1 roll exch L 4 2 roll L L cp}bd/RRCC{/r ed/y1 ed/x1 ed/y0 ed/x0
ed x0 x1 add 2 div y0 M x1 y0 x1 y1 r arcto 4{pop}repeat x1 y1 x0 y1
r arcto 4{pop}repeat x0 y1 x0 y0 r arcto 4{pop}repeat x0 y0 x1 y0 r
arcto 4{pop}repeat cp}bd/RR{2 copy 0 eq exch 0 eq or{pop pop RRect}{2
copy eq{pop RRCC}{m cm pop/y2 ed/x2 ed/ys y2 x2 div 1 max def/xs x2
y2 div 1 max def/y1 exch ys div def/x1 exch xs div def/y0 exch ys div
def/x0 exch xs div def/r2 x2 y2 min def xs ys scale x0 x1 add 2 div
y0 M x1 y0 x1 y1 r2 arcto 4{pop}repeat x1 y1 x0 y1 r2 arcto 4{pop}repeat
x0 y1 x0 y0 r2 arcto 4{pop}repeat x0 y0 x1 y0 r2 arcto 4{pop}repeat
m sm cp}ifelse}ifelse}bd/PP{{rlt}repeat}bd/OB{gs 0 ne{7 3 roll/y ed
/x ed x y translate ULesc rotate x neg y neg translate x y 7 -3 roll}if
sc B fill gr}bd/B{M/dy ed/dx ed dx 0 rlt 0 dy rlt dx neg 0 rlt cp}bd
/CB{B clip n}bd/ErrHandler{errordict dup maxlength exch length gt
dup{errordict begin}if/errhelpdict 12 dict def errhelpdict begin/stackunderflow(operand stack underflow)def
/undefined(this name is not defined in a dictionary)def/VMerror(you have used up all the printer's memory)def
/typecheck(operator was expecting a different type of operand)def
/ioerror(input/output error occured)def end{end}if errordict begin
/handleerror{$error begin newerror{/newerror false def showpage 72
72 scale/x .25 def/y 9.6 def/Helvetica findfont .2 scalefont setfont
x y moveto(Offending Command = )show/command load{dup type/stringtype
ne{(max err string)cvs}if show}exec/y y .2 sub def x y moveto(Error = )show
errorname{dup type dup( max err string )cvs show( : )show/stringtype
ne{( max err string )cvs}if show}exec errordict begin errhelpdict errorname
known{x 1 add y .2 sub moveto errhelpdict errorname get show}if end
/y y .4 sub def x y moveto(Stack =)show ostack{/y y .2 sub def x 1
add y moveto dup type/stringtype ne{( max err string )cvs}if show}forall
showpage}if end}def end}bd end
/SVDoc save def
Win35Dict begin
ErrHandler
statusdict begin 0 setjobtimeout end
statusdict begin statusdict /jobname (Microsoft Word ภาษาไทย - FI...) put end
/oldDictCnt countdictstack def {}stopped 
{ countdictstack oldDictCnt lt { Win35Dict begin } 
{1 1 countdictstack oldDictCnt sub {pop end } for } ifelse } if 
/oldDictCnt countdictstack def {a4
}stopped 
{ countdictstack oldDictCnt lt { Win35Dict begin } 
{1 1 countdictstack oldDictCnt sub {pop end } for } ifelse } if 
[
{mark 1.0 1.0 .98 .9 .82 .68 .56 .48 .28 .1 .06 .0 counttomark dup 3 add -1 roll exch 2 sub mul dup floor cvi dup 3 1 roll sub exch dup  3 add index exch 2 add index dup 4 1 roll sub mul add counttomark 1 add 1 roll  cleartomark } bind
/exec load currenttransfer /exec load] cvx settransfer
gsave %matrix defaultmatrix setmatrix
90 rotate 4 72 mul .55 -72 mul moveto /Times-Roman findfont
20 scalefont setfont 0.3 setgray (cmp-lg/9411027   7 Oct 1995) show grestore
%%PageResources: (atend)
SS
0 0 25 31 776 1169 300 SM
32 0 0 58 58 0 0 0 56 /Times-Bold /font29 ANSIFont font
0 0 0 fC
295 207 1660 (CLASSIFIER ASSIGNMENT BY CORPUS-BASED APPROACH) 1660 SB
32 0 0 50 50 0 0 0 48 /Times-Bold /font29 ANSIFont font
235 389 561 (Virach Sornlertlamvanich) 561 SB
835 389 552 (        Wantanee Pantachat) 552 SB
1585 389 431 (Surapant Meknavin) 431 SB
32 0 0 42 42 0 0 0 38 /Times-Roman /font32 ANSIFont font
722 503 807 (Linguistics and Knowledge Science Laboratory) 807 SB
657 553 937 (National Electronics and Computer Technology Center) 937 SB
648 603 955 (National Science and Technology Development Agency) 955 SB
700 653 851 (Ministry of Science Technology and Environment) 851 SB
825 703 601 (22nd Gypsum Metropolitan Tower,) 601 SB
709 753 833 (539/2 Sriayudhya Rd., Bangkok 10400, Thailand) 833 SB
735 803 780 ({virach,wantanee,surapan}@nwg.nectec.or.th) 780 SB
32 0 0 46 46 0 0 0 44 /Times-Bold /font29 ANSIFont font
550 1000 170 (Abstract) 170 SB
32 0 0 42 42 0 0 0 38 /Times-Roman /font32 ANSIFont font
165 1109 86 (This ) 102 SB
267 1109 105 (paper ) 121 SB
388 1109 149 (presents ) 165 SB
553 1109 51 (an ) 67 SB
620 1109 176 (algorithm ) 192 SB
812 1109 60 (for ) 76 SB
888 1109 162 (selecting ) 178 SB
1066 1109 40 (an) 40 SB
165 1159 204 (appropriate ) 226 SB
391 1159 164 (classifier ) 186 SB
577 1159 97 (word ) 119 SB
696 1159 60 (for ) 82 SB
778 1159 30 (a ) 52 SB
830 1159 106 (noun. ) 129 SB
959 1159 46 (In ) 69 SB
1028 1159 78 (Thai) 78 SB
165 1209 619 (language, it frequently happens that ) 620 SB
785 1209 96 (there ) 97 SB
882 1209 39 (is ) 40 SB
922 1209 184 (fluctuation) 184 SB
165 1259 44 (in ) 50 SB
215 1259 63 (the ) 69 SB
284 1259 122 (choice ) 128 SB
412 1259 46 (of ) 52 SB
464 1259 164 (classifier ) 170 SB
634 1259 60 (for ) 66 SB
700 1259 30 (a ) 37 SB
737 1259 105 (given ) 112 SB
849 1259 155 (concrete ) 162 SB
1011 1259 95 (noun,) 95 SB
165 1309 86 (both ) 99 SB
264 1309 93 (from ) 106 SB
370 1309 63 (the ) 77 SB
447 1309 98 (point ) 112 SB
559 1309 46 (of ) 60 SB
619 1309 93 (view ) 107 SB
726 1309 46 (of ) 60 SB
786 1309 63 (the ) 77 SB
863 1309 114 (whole ) 128 SB
991 1309 115 (speech) 115 SB
165 1359 204 (community ) 205 SB
370 1359 72 (and ) 74 SB
444 1359 183 (individual ) 185 SB
629 1359 167 (speakers. ) 169 SB
798 1359 180 (Basically, ) 182 SB
980 1359 96 (there ) 98 SB
1078 1359 28 (is) 28 SB
165 1409 53 (no ) 56 SB
221 1409 101 (exact ) 104 SB
325 1409 77 (rule ) 80 SB
405 1409 60 (for ) 63 SB
468 1409 164 (classifier ) 167 SB
635 1409 173 (selection. ) 176 SB
811 1409 57 (As ) 60 SB
871 1409 58 (far ) 62 SB
933 1409 46 (as ) 50 SB
983 1409 60 (we ) 64 SB
1047 1409 59 (can) 59 SB
165 1459 53 (do ) 56 SB
221 1459 44 (in ) 47 SB
268 1459 63 (the ) 66 SB
334 1459 187 (rule-based ) 190 SB
524 1459 166 (approach ) 170 SB
694 1459 39 (is ) 43 SB
737 1459 44 (to ) 48 SB
785 1459 84 (give ) 88 SB
873 1459 30 (a ) 34 SB
907 1459 129 (default ) 133 SB
1040 1459 66 (rule) 66 SB
165 1509 44 (to ) 54 SB
219 1509 84 (pick ) 94 SB
313 1509 53 (up ) 63 SB
376 1509 30 (a ) 40 SB
416 1509 252 (corresponding ) 263 SB
679 1509 164 (classifier ) 175 SB
854 1509 46 (of ) 57 SB
911 1509 89 (each ) 100 SB
1011 1509 95 (noun.) 95 SB
165 1559 843 (Registration of classifier for each noun is limited ) 844 SB
1009 1559 44 (to ) 45 SB
1054 1559 52 (the) 52 SB
165 1609 84 (type ) 85 SB
250 1609 46 (of ) 47 SB
297 1609 77 (unit ) 78 SB
375 1609 164 (classifier ) 165 SB
540 1609 145 (because ) 146 SB
686 1609 98 (other ) 99 SB
785 1609 100 (types ) 101 SB
886 1609 63 (are ) 64 SB
950 1609 93 (open ) 95 SB
1045 1609 61 (due) 61 SB
165 1659 44 (to ) 63 SB
228 1659 63 (the ) 82 SB
310 1659 157 (meaning ) 176 SB
486 1659 46 (of ) 65 SB
551 1659 262 (representation. ) 282 SB
833 1659 70 (We ) 90 SB
923 1659 144 (propose ) 164 SB
1087 1659 19 (a) 19 SB
165 1709 233 (corpus-based ) 275 SB
440 1709 138 (method ) 180 SB
620 1709 226 (\(Biber,1993; ) 269 SB
889 1709 217 (Nagao,1993;) 217 SB
165 1759 247 (Smadja,1993\) ) 282 SB
447 1759 114 (which ) 150 SB
597 1759 171 (generates ) 207 SB
804 1759 104 (Noun ) 140 SB
944 1759 162 (Classifier) 162 SB
165 1809 226 (Associations ) 243 SB
408 1809 127 (\(NCA\) ) 144 SB
552 1809 44 (to ) 61 SB
613 1809 178 (overcome ) 195 SB
808 1809 63 (the ) 80 SB
888 1809 168 (problems ) 185 SB
1073 1809 33 (in) 33 SB
165 1859 164 (classifier ) 181 SB
346 1859 201 (assignment ) 218 SB
564 1859 72 (and ) 89 SB
653 1859 162 (semantic ) 180 SB
833 1859 220 (construction ) 238 SB
1071 1859 35 (of) 35 SB
165 1909 95 (noun ) 101 SB
266 1909 132 (phrase. ) 138 SB
404 1909 77 (The ) 83 SB
487 1909 99 (NCA ) 105 SB
592 1909 39 (is ) 45 SB
637 1909 134 (created ) 140 SB
777 1909 205 (statistically ) 211 SB
988 1909 93 (from ) 99 SB
1087 1909 19 (a) 19 SB
165 1959 96 (large ) 100 SB
265 1959 123 (corpus ) 127 SB
392 1959 72 (and ) 76 SB
468 1959 215 (recomposed ) 219 SB
687 1959 107 (under ) 111 SB
798 1959 143 (concept ) 148 SB
946 1959 160 (hierarchy) 160 SB
165 2009 705 (constraints and frequency of occurrences.) 705 SB
165 2109 856 (Keywords: Thai language, classifier, corpus-based) 856 SB
165 2159 768 (method, Noun Classifier Associations \(NCA\)) 768 SB
32 0 0 46 46 0 0 0 44 /Times-Bold /font29 ANSIFont font
165 2306 298 (1. Introduction) 298 SB
32 0 0 42 42 0 0 0 38 /Times-Roman /font32 ANSIFont font
165 2427 41 (A ) 45 SB
210 2427 164 (classifier ) 168 SB
378 2427 67 (has ) 71 SB
449 2427 30 (a ) 34 SB
483 2427 190 (significant ) 194 SB
677 2427 67 (use ) 71 SB
748 2427 44 (in ) 48 SB
796 2427 89 (Thai ) 93 SB
889 2427 164 (language ) 168 SB
1057 2427 49 (for) 49 SB
165 2477 220 (construction ) 236 SB
401 2477 46 (of ) 62 SB
463 2477 95 (noun ) 112 SB
575 2477 46 (or ) 63 SB
638 2477 86 (verb ) 103 SB
741 2477 44 (to ) 61 SB
802 2477 137 (express ) 154 SB
956 2477 150 (quantity,) 150 SB
165 2527 258 (determination, ) 261 SB
426 2527 162 (pronoun, ) 166 SB
592 2527 72 (etc. ) 76 SB
668 2527 60 (By ) 64 SB
732 2527 58 (far ) 62 SB
794 2527 63 (the ) 67 SB
861 2527 93 (most ) 97 SB
958 2527 148 (common) 148 SB
165 2577 67 (use ) 71 SB
236 2577 46 (of ) 51 SB
287 2577 191 (classifiers, ) 196 SB
483 2577 167 (however, ) 172 SB
655 2577 39 (is ) 44 SB
699 2577 44 (in ) 49 SB
748 2577 250 (enumerations, ) 255 SB
1003 2577 103 (where) 103 SB
165 2627 63 (the ) 104 SB
269 2627 180 (classifiers ) 221 SB
490 2627 121 (follow ) 162 SB
652 2627 166 (numerals ) 208 SB
860 2627 72 (and ) 114 SB
974 2627 132 (precede) 132 SB
165 2677 267 (demonstratives ) 270 SB
435 2677 228 (\(Noss,1964\). ) 231 SB
666 2677 74 (Not ) 77 SB
743 2677 54 (all ) 57 SB
800 2677 100 (types ) 103 SB
903 2677 46 (of ) 50 SB
953 2677 153 (classifier) 153 SB
165 2727 91 (have ) 108 SB
273 2727 30 (a ) 47 SB
320 2727 211 (relationship ) 228 SB
548 2727 86 (with ) 103 SB
651 2727 95 (noun ) 112 SB
763 2727 46 (or ) 63 SB
826 2727 86 (verb ) 103 SB
929 2727 46 (as ) 63 SB
992 2727 30 (a ) 48 SB
1040 2727 66 (unit) 66 SB
165 2777 252 (classifier does.) 252 SB
315 2827 41 (A ) 45 SB
32 0 0 42 42 0 0 0 39 /Times-Italic /font31 ANSIFont font
360 2826 77 (unit ) 81 SB
441 2826 155 (classifier) 155 SB
32 0 0 42 42 0 0 0 38 /Times-Roman /font32 ANSIFont font
596 2827 11 ( ) 15 SB
611 2827 39 (is ) 43 SB
654 2827 72 (any ) 76 SB
730 2827 164 (classifier ) 168 SB
898 2827 114 (which ) 118 SB
1016 2827 67 (has ) 71 SB
1087 2827 19 (a) 19 SB
165 2879 129 (special ) 133 SB
298 2879 211 (relationship ) 215 SB
513 2879 86 (with ) 91 SB
604 2879 72 (one ) 77 SB
681 2879 46 (or ) 51 SB
732 2879 98 (more ) 103 SB
835 2879 155 (concrete ) 160 SB
995 2879 111 (nouns.) 111 SB
165 2929 69 (For ) 74 SB
239 2929 166 (example, ) 171 SB
410 2929 44 (to ) 49 SB
459 2929 188 (enumerate ) 193 SB
652 2929 166 (members ) 171 SB
823 2929 46 (of ) 51 SB
874 2929 63 (the ) 69 SB
943 2929 93 (class ) 99 SB
1042 2929 46 (of ) 52 SB
1094 2929 12 (/) 12 SB
165 2979 607 (rya/ 'boats', the unit classifier /lam/ ) 608 SB
773 2979 39 (is ) 40 SB
813 2979 148 (selected ) 149 SB
962 2979 46 (as ) 47 SB
1009 2979 44 (in ) 45 SB
1054 2979 52 (the) 52 SB
165 3029 236 (phrase below:) 236 SB

/GreNewFont{10 dict dup 3 1 roll def dup begin 6 1 roll/FontType 3
def/FontMatrix exch def/FontBBox exch def/FontInfo 2 dict def FontInfo
/UnderlinePosition 3 -1 roll put FontInfo/UnderlineThickness 3 -1
roll put/Encoding 256 array def 0 1 255{Encoding exch/.notdef put}for
/CharProcs 256 dict def CharProcs/.notdef{}put/Metrics 256 dict def
Metrics/.notdef 3 -1 roll put/BuildChar{/char exch def/fontdict exch
def/charname fontdict/Encoding get char get def fontdict/Metrics get
charname get aload pop setcachedevice fontdict begin Encoding char
get CharProcs exch get end exec}def end definefont pop}def/AddChar{begin
Encoding 3 1 roll put CharProcs 3 1 roll put Metrics 3 1 roll put end}def
/MSTT31c146 [50.0 0 0 0 0 0] 20 -40 [-50.0 -50.0 50.0 50.0] [1 50 div 0 0 1 50 div 0 0] /MSTT31c146 GreNewFont

32 0 0 50 50 0 0 1 47 /MSTT31c146 font

/Ge0 [8.0 0.0 1.0 0.0 7.0 20.0]
/Ge0 {
    6 20 true [1 0 0 -1 -1.0 20.0] {<10307070707070707070707070707844c4c44438>} imagemask 
  }
  224 /Ge0 MSTT31c146 AddChar
/Gc3 [17.0 0.0 1.0 0.0 14.0 20.0]
/Gc3 {
    13 20 true [1 0 0 -1 -1.0 20.0] {<1e083ff823f060206000fc000fc001e000e000e000e000e000e000e000e007e008e008e018c00780
>} imagemask 
  }
  195 /Gc3 MSTT31c146 AddChar
/Gd7 [0.0 0.0 -17.0 23.0 -1.0 34.0]
/Gd7 {
    16 11 true [1 0 0 -1 17.0 34.0] {<00c600c600c607e61ffe383e600e4006ffe2803f0003>} imagemask 
  }
  215 /Gd7 MSTT31c146 AddChar
/G20 [11.0 0.0 0.0 0.0 0.0 0.0]
/G20 {
} 
  32 /G20 MSTT31c146 AddChar
/Gcd [20.0 0.0 1.0 0.0 17.0 20.0]
/Gcd {
    16 20 true [1 0 0 -1 -1.0 20.0] {<0fe0303c600e4006ce071d073887308739873f073e073807380738073807380738071c0e0e3c03e0
>} imagemask 
  }
  205 /Gcd MSTT31c146 AddChar
/Gcb [23.0 0.0 2.0 0.0 20.0 20.0]
/Gcb {
    18 20 true [1 0 0 -1 -2.0 20.0] {<780380dc06c08e0c408e0c40de06c07e0f800e1f800e33800e63800e43800e83800f03800f03800e
03800e03800e03800e03800e03800e03800e0380>} imagemask 
  }
  203 /Gcb MSTT31c146 AddChar
/Gb9 [23.0 0.0 2.0 0.0 20.0 20.0]
/Gb9 {
    18 20 true [1 0 0 -1 -2.0 20.0] {<780080dc03808e03808e0380fe03802e03800e03800e03800e03800e03800e03800e03800e03800e
03800e3f000ec7800f88400f08400e0cc00e0780>} imagemask 
  }
  185 /Gb9 MSTT31c146 AddChar
/Gd6 [0.0 0.0 -18.0 23.0 -1.0 34.0]
/Gd6 {
    17 11 true [1 0 0 -1 18.0 34.0] {<001f000011800031800ff1801ffb80301f00600600c00200fff300000f00000100>} imagemask 
  }
  214 /Gd6 MSTT31c146 AddChar
/Ge8 [0.0 0.0 -7.0 36.0 -3.0 42.0]
/Ge8 {
    4 6 true [1 0 0 -1 7.0 42.0] {<f0f0f0e06060>} imagemask 
  }
  232 /Ge8 MSTT31c146 AddChar
/Ga7 [16.0 0.0 2.0 0.0 13.0 20.0]
/Ga7 {
    11 20 true [1 0 0 -1 -2.0 20.0] {<07800dc008e008e00ce007e000e000e080e080e0c0e060e030e018e00ce00fe007e003e001e000e0
>} imagemask 
  }
  167 /Ga7 MSTT31c146 AddChar
/Gc5 [20.0 0.0 1.0 0.0 17.0 20.0]
/Gc5 {
    16 20 true [1 0 0 -1 -1.0 20.0] {<07e0383c400e800703c70fe71c1f1c0f3807380738073807380738073f0739873087308719870f07
>} imagemask 
  }
  197 /Gc5 MSTT31c146 AddChar
/Ged [0.0 0.0 -11.0 22.0 -3.0 29.0]
/Ged {
    8 7 true [1 0 0 -1 11.0 29.0] {<3c7ec3c3c37e3c>} imagemask 
  }
  237 /Ged MSTT31c146 AddChar
/Gd2 [14.0 0.0 1.0 0.0 12.0 20.0]
/Gd2 {
    11 20 true [1 0 0 -1 -1.0 20.0] {<1f0077c0c1c080e000e000e000e000e000e000e000e000e000e000e000e000e000e000e000e000e0
>} imagemask 
  }
  210 /Gd2 MSTT31c146 AddChar

1360 1050 36 (\340\303\327 ) 25 SB
1385 1050 143 (\315      \313\271\326\350 ) 132 SB
1517 1050 102 (\247     \305\355 ) 91 SB
1608 1050 14 (\322) 15 SB
32 0 0 42 42 0 0 0 38 /Times-Roman /font32 ANSIFont font
1360 1120 314 (/rya    nung    lam/) 314 SB
1360 1170 332 (boat   one    <boat>) 332 SB
1360 1220 172 ('one boat'.) 172 SB
1210 1320 107 (Other ) 122 SB
1332 1320 84 (than ) 99 SB
1431 1320 63 (the ) 79 SB
1510 1320 77 (unit ) 93 SB
1603 1320 175 (classifier, ) 191 SB
1794 1320 96 (there ) 112 SB
1906 1320 63 (are ) 79 SB
1985 1320 166 (collective) 166 SB
1210 1370 175 (classifier, ) 188 SB
1398 1370 120 (metric ) 133 SB
1531 1370 175 (classifier, ) 188 SB
1719 1370 180 (frequency ) 193 SB
1912 1370 164 (classifier ) 178 SB
2090 1370 61 (and) 61 SB
1210 1420 281 (verbal classifier.) 281 SB
1360 1470 41 (A ) 46 SB
32 0 0 42 42 0 0 0 39 /Times-Italic /font31 ANSIFont font
1406 1469 175 (collective ) 180 SB
1586 1469 155 (classifier) 155 SB
32 0 0 42 42 0 0 0 38 /Times-Roman /font32 ANSIFont font
1741 1470 11 ( ) 16 SB
1757 1470 39 (is ) 44 SB
1801 1470 72 (any ) 77 SB
1878 1470 164 (classifier ) 170 SB
2048 1470 103 (which) 103 SB
1210 1528 115 (shows ) 118 SB
1328 1528 136 (general ) 139 SB
1467 1528 109 (group ) 112 SB
1579 1528 46 (or ) 49 SB
1628 1528 58 (set ) 61 SB
1689 1528 46 (of ) 49 SB
1738 1528 95 (mass ) 98 SB
1836 1528 122 (nouns, ) 125 SB
32 0 0 50 50 0 0 1 47 /MSTT31c146 font

/Ga1 [20.0 0.0 1.0 0.0 17.0 20.0]
/Ga1 {
    16 20 true [1 0 0 -1 -1.0 20.0] {<07f01e7c301e600e4007ff0738073807380738073807380738073807380738073807380738063804
>} imagemask 
  }
  161 /Ga1 MSTT31c146 AddChar
/Gca [22.0 0.0 1.0 0.0 21.0 22.0]
/Gca {
    20 22 true [1 0 0 -1 -1.0 22.0] {<00007000007003f0700ffc60181fe02007c0409f80c3fb80071f800e07801c03801c03801c03801c
03801c03801c03801f03801c838018c3801843800c8380078380>} imagemask 
  }
  202 /Gca MSTT31c146 AddChar
/Gbd [19.0 0.0 1.0 0.0 17.0 33.0]
/Gbd {
    16 33 true [1 0 0 -1 -1.0 33.0] {<00010003000700070007000700070007000700070007000700073c077607e207e207fe07ec07e007
e087e187e3c7e247e667ec37e817f81ff00fe007e007e006e004>} imagemask 
  }
  189 /Gbd MSTT31c146 AddChar
/Gd9 [0.0 0.0 -14.0 -10.0 -3.0 -2.0]
/Gd9 {
    11 8 true [1 0 0 -1 14.0 -2.0] {<78e08ce08ce08ce0fce00ce00fe007c0>} imagemask 
  }
  217 /Gd9 MSTT31c146 AddChar

1961 1519 54 (\271\241 ) 57 SB
2018 1519 69 (\312\315\247 ) 72 SB
2090 1519 30 (\275\331 ) 19 SB
2109 1519 27 (\247 ) 30 SB
32 0 0 42 42 0 0 0 38 /Times-Roman /font32 ANSIFont font
2139 1528 12 (/) 12 SB
1210 1589 74 (nok ) 86 SB
1296 1589 11 ( ) 23 SB
1319 1589 111 (soong ) 123 SB
1442 1589 11 ( ) 23 SB
1465 1589 100 (fung/ ) 112 SB
1577 1589 11 ( ) 23 SB
1600 1589 82 ('two ) 94 SB
1694 1589 114 (flocks ) 126 SB
1820 1589 46 (of ) 58 SB
1878 1589 98 (bird'. ) 111 SB
1989 1589 41 (A ) 54 SB
32 0 0 42 42 0 0 0 39 /Times-Italic /font31 ANSIFont font
2043 1588 108 (metric) 108 SB
1210 1640 155 (classifier) 155 SB
32 0 0 42 42 0 0 0 38 /Times-Roman /font32 ANSIFont font
1365 1641 11 ( ) 49 SB
1414 1641 39 (is ) 77 SB
1491 1641 72 (any ) 111 SB
1602 1641 164 (classifier ) 203 SB
1805 1641 114 (which ) 153 SB
1958 1641 121 (occurs ) 160 SB
2118 1641 33 (in) 33 SB
1210 1693 239 (enumerations ) 242 SB
1452 1693 75 (that ) 78 SB
1530 1693 133 (modify ) 136 SB
1666 1693 183 (predicates ) 186 SB
1852 1693 46 (as ) 50 SB
1902 1693 84 (well ) 88 SB
1990 1693 46 (as ) 50 SB
2040 1693 111 (nouns,) 111 SB
32 0 0 50 50 0 0 1 47 /MSTT31c146 font

/Ge9 [0.0 0.0 -8.0 36.0 3.0 46.0]
/Ge9 {
    11 10 true [1 0 0 -1 8.0 46.0] {<00607860cc608c404cc03cc019801b007e00f000>} imagemask 
  }
  233 /Ge9 MSTT31c146 AddChar
/Gc1 [21.0 0.0 1.0 0.0 18.0 20.0]
/Gc1 {
    17 20 true [1 0 0 -1 -1.0 20.0] {<780080cc0180860380860380ce03807e03800e03800e03800e03800e03800e03800e03800e03800e
03803f83804ec3804e63804e3b807c1f001c0e00>} imagemask 
  }
  193 /Gc1 MSTT31c146 AddChar
/Ge1 [19.0 0.0 1.0 0.0 18.0 20.0]
/Ge1 {
    17 20 true [1 0 0 -1 -1.0 20.0] {<100400300c00701c00701c00701c00701c00701c00701c00701c00701c00701c00701c00701c0070
1c00781e00441100c42100c42180441100380e00>} imagemask 
  }
  225 /Ge1 MSTT31c146 AddChar
/G8c [0.0 0.0 -12.0 22.0 1.0 35.0]
/G8c {
    13 13 true [1 0 0 -1 12.0 35.0] {<001810186c18c6188638c6307e301e600ce00dc01f00fc00e000>} imagemask 
  }
  140 /G8c MSTT31c146 AddChar
/Gc7 [18.0 0.0 2.0 0.0 15.0 20.0]
/Gc7 {
    13 20 true [1 0 0 -1 -2.0 20.0] {<0f803fe06070c038803800380038003800380038003800380038003801f8037802380238033001e0
>} imagemask 
  }
  199 /Gc7 MSTT31c146 AddChar

1210 1740 34 (\271\355\351 ) 23 SB
1233 1740 14 (\322) 15 SB
1248 1740 11 ( ) 13 SB
1261 1740 68 (\312\322\301 ) 70 SB
1331 1740 50 (\341\241\214 ) 39 SB
1370 1740 29 (\307 ) 31 SB
32 0 0 42 42 0 0 0 38 /Times-Roman /font32 ANSIFont font
1401 1749 96 (/nam ) 98 SB
1499 1749 11 ( ) 13 SB
1512 1749 98 (saam ) 100 SB
1612 1749 11 ( ) 13 SB
1625 1749 112 (kaew/ ) 114 SB
1739 1749 11 ( ) 13 SB
1752 1749 104 ('three ) 106 SB
1858 1749 130 (glasses ) 132 SB
1990 1749 46 (of ) 48 SB
2038 1749 113 (water'.) 113 SB
1210 1810 41 (A ) 41 SB
32 0 0 42 42 0 0 0 39 /Times-Italic /font31 ANSIFont font
1251 1809 333 (frequency classifier) 333 SB
32 0 0 42 42 0 0 0 38 /Times-Roman /font32 ANSIFont font
1584 1810 11 ( ) 12 SB
1596 1810 39 (is ) 40 SB
1636 1810 72 (any ) 73 SB
1709 1810 164 (classifier ) 165 SB
1874 1810 114 (which ) 115 SB
1989 1810 39 (is ) 40 SB
2029 1810 88 (used ) 89 SB
2118 1810 33 (to) 33 SB
1210 1868 137 (express ) 140 SB
1350 1868 63 (the ) 66 SB
1416 1868 180 (frequency ) 183 SB
1599 1868 46 (of ) 49 SB
1648 1868 103 (event ) 106 SB
1754 1868 75 (that ) 78 SB
1832 1868 132 (occurs, ) 136 SB
32 0 0 50 50 0 0 1 47 /MSTT31c146 font

/Gba [22.0 0.0 1.0 0.0 20.0 20.0]
/Gba {
    19 20 true [1 0 0 -1 -1.0 20.0] {<3c00e07e00e0c700e08700e04f00e07f00e00700e00700e00700e00700e00700e00700e00700e007
00e00700e00700e00700e00700e00701e01fff80>} imagemask 
  }
  186 /Gba MSTT31c146 AddChar
/Gd4 [0.0 0.0 -17.0 24.0 -1.0 32.0]
/Gd4 {
    16 8 true [1 0 0 -1 17.0 32.0] {<07e01ff8381c60064002ffe3003f0003>} imagemask 
  }
  212 /Gd4 MSTT31c146 AddChar
/Gd5 [0.0 0.0 -19.0 23.0 -2.0 33.0]
/Gd5 {
    17 10 true [1 0 0 -1 19.0 33.0] {<00070000070007f700181f00200700400300ffc100c07f80000380000080>} imagemask 
  }
  213 /Gd5 MSTT31c146 AddChar

1968 1859 33 (\272\324 ) 22 SB
1990 1859 34 (\271 ) 38 SB
2028 1859 33 (\312\325\350 ) 22 SB
2050 1859 11 ( ) 15 SB
2065 1859 70 (\303\315\272 ) 74 SB
32 0 0 42 42 0 0 0 38 /Times-Roman /font32 ANSIFont font
2139 1868 12 (/) 12 SB
1210 1929 65 (bin ) 68 SB
1278 1929 11 ( ) 15 SB
1293 1929 51 (sii ) 55 SB
1348 1929 11 ( ) 15 SB
1363 1929 100 (roob/ ) 104 SB
1467 1929 11 ( ) 15 SB
1482 1929 66 ('fly ) 70 SB
1552 1929 81 (four ) 85 SB
1637 1929 144 (rounds'. ) 148 SB
1785 1929 41 (A ) 45 SB
32 0 0 42 42 0 0 0 39 /Times-Italic /font31 ANSIFont font
1830 1928 119 (verbal ) 123 SB
1953 1928 155 (classifier) 155 SB
32 0 0 42 42 0 0 0 38 /Times-Roman /font32 ANSIFont font
2108 1929 11 ( ) 15 SB
2123 1929 28 (is) 28 SB
1210 1981 72 (any ) 73 SB
1283 1981 164 (classifier ) 165 SB
1448 1981 114 (which ) 115 SB
1563 1981 39 (is ) 40 SB
1603 1981 138 (derived ) 139 SB
1742 1981 93 (from ) 94 SB
1836 1981 30 (a ) 31 SB
1867 1981 86 (verb ) 88 SB
1955 1981 72 (and ) 74 SB
2029 1981 122 (usually) 122 SB
1210 2037 88 (used ) 91 SB
1301 2037 44 (in ) 47 SB
1348 2037 220 (construction ) 223 SB
1571 2037 86 (with ) 89 SB
1660 2037 95 (mass ) 98 SB
1758 2037 122 (nouns, ) 126 SB
32 0 0 50 50 0 0 1 47 /MSTT31c146 font

/Gd0 [18.0 0.0 1.0 0.0 15.0 19.0]
/Gd0 {
    14 19 true [1 0 0 -1 -1.0 19.0] {<7c0cc60cc208c218c6707fe01f800000000000000000100c7c0cc60cc218c238e6707fe01f00>} imagemask 
  }
  208 /Gd0 MSTT31c146 AddChar
/Gb4 [20.0 0.0 2.0 0.0 17.0 20.0]
/Gb4 {
    15 20 true [1 0 0 -1 -2.0 20.0] {<07e01ff8383c601cc00ec00e878e844ec44ec44e66ce738e330e3e0e3c0e380e380e380e380e380e
>} imagemask 
  }
  180 /Gb4 MSTT31c146 AddChar
/Gc9 [22.0 0.0 1.0 0.0 19.0 20.0]
/Gc9 {
    18 20 true [1 0 0 -1 -1.0 20.0] {<780040cc01c08e01c08e01c07e01c00e01c00e01c00e41c00fe1c00f31c00f21c00f01c00f87c00e
fdc00e01c00e01c00e01c00e01c00e03803fff00>} imagemask 
  }
  201 /Gc9 MSTT31c146 AddChar

1884 2028 122 (\241\303\320\264\322\311 ) 126 SB
2010 2028 34 (\313\214 ) 23 SB
2033 2028 25 (\322 ) 29 SB
2062 2028 32 (\301\214 ) 21 SB
2083 2028 52 (\307\271 ) 56 SB
32 0 0 42 42 0 0 0 38 /Times-Roman /font32 ANSIFont font
2139 2037 12 (/) 12 SB
1210 2098 695 (kradaad  haa  muan/  'five rolls of paper'.) 695 SB
1360 2148 77 (The ) 84 SB
1444 2148 77 (unit ) 85 SB
1529 2148 164 (classifier ) 172 SB
1701 2148 67 (has ) 75 SB
1776 2148 30 (a ) 38 SB
1814 2148 129 (special ) 137 SB
1951 2148 200 (relationship) 200 SB
1210 2198 86 (with ) 101 SB
1311 2198 155 (concrete ) 170 SB
1481 2198 106 (noun. ) 121 SB
1602 2198 77 (The ) 92 SB
1694 2198 150 (member ) 165 SB
1859 2198 46 (of ) 61 SB
1920 2198 72 (this ) 87 SB
2007 2198 93 (class ) 109 SB
2116 2198 35 (of) 35 SB
1210 2248 164 (classifier ) 174 SB
1384 2248 39 (is ) 49 SB
1433 2248 119 (closed ) 129 SB
1562 2248 60 (for ) 70 SB
1632 2248 89 (each ) 99 SB
1731 2248 106 (noun. ) 116 SB
1847 2248 97 (Most ) 107 SB
1954 2248 46 (of ) 57 SB
2011 2248 63 (the ) 74 SB
2085 2248 66 (unit) 66 SB
1210 2298 180 (classifiers ) 184 SB
1394 2298 63 (are ) 67 SB
1461 2298 88 (used ) 93 SB
1554 2298 86 (with ) 91 SB
1645 2298 30 (a ) 35 SB
1680 2298 96 (great ) 101 SB
1781 2298 105 (many ) 110 SB
1891 2298 155 (concrete ) 160 SB
2051 2298 100 (nouns) 100 SB
1210 2348 46 (of ) 51 SB
1261 2348 86 (very ) 91 SB
1352 2348 157 (different ) 162 SB
1514 2348 168 (meaning, ) 173 SB
1687 2348 65 (but ) 71 SB
1758 2348 74 (few ) 80 SB
1838 2348 63 (are ) 69 SB
1907 2348 169 (restricted ) 175 SB
2082 2348 44 (to ) 50 SB
2132 2348 19 (a) 19 SB
1210 2398 785 (single noun. Except for the unit classifier, the ) 786 SB
1996 2398 155 (members) 155 SB
1210 2448 46 (of ) 68 SB
1278 2448 164 (classifier ) 186 SB
1464 2448 60 (for ) 82 SB
1546 2448 30 (a ) 52 SB
1598 2448 95 (noun ) 117 SB
1715 2448 46 (or ) 68 SB
1783 2448 167 (predicate ) 189 SB
1972 2448 63 (are ) 86 SB
2058 2448 93 (open.) 93 SB
1210 2498 188 (Especially ) 201 SB
1411 2498 60 (for ) 74 SB
1485 2498 63 (the ) 77 SB
1562 2498 120 (metric ) 134 SB
1696 2498 175 (classifier, ) 189 SB
1885 2498 63 (the ) 77 SB
1962 2498 140 (number ) 154 SB
2116 2498 35 (of) 35 SB
1210 2548 180 (classifiers ) 192 SB
1402 2548 60 (for ) 72 SB
1474 2548 150 (numeral ) 163 SB
1637 2548 191 (expression ) 204 SB
1841 2548 46 (of ) 59 SB
1900 2548 161 (distance, ) 174 SB
2074 2548 77 (size,) 77 SB
1210 2598 616 (weight, container and value is large.) 616 SB
1360 2648 77 (The ) 80 SB
1440 2648 67 (use ) 71 SB
1511 2648 46 (of ) 50 SB
1561 2648 164 (classifier ) 168 SB
1729 2648 44 (in ) 48 SB
1777 2648 89 (Thai ) 93 SB
1870 2648 39 (is ) 43 SB
1913 2648 65 (not ) 69 SB
1982 2648 132 (limited ) 136 SB
2118 2648 33 (to) 33 SB
1210 2698 63 (the ) 82 SB
1292 2698 150 (numeral ) 169 SB
1461 2698 191 (expression ) 211 SB
1672 2698 65 (but ) 85 SB
1757 2698 39 (is ) 59 SB
1816 2698 164 (extended ) 184 SB
2000 2698 44 (to ) 64 SB
2064 2698 87 (other) 87 SB
1210 2748 207 (expressions ) 221 SB
1431 2748 88 (such ) 102 SB
1533 2748 46 (as ) 60 SB
1593 2748 142 (ordinal, ) 157 SB
1750 2748 258 (determination, ) 273 SB
2023 2748 128 (relative) 128 SB
1210 2798 162 (pronoun, ) 172 SB
1382 2798 162 (pronoun, ) 172 SB
1554 2798 72 (etc. ) 82 SB
1636 2798 77 (The ) 88 SB
1724 2798 106 (detail ) 117 SB
1841 2798 46 (of ) 57 SB
1898 2798 89 (each ) 100 SB
1998 2798 153 (classifier) 153 SB
1210 2848 655 (phrase is described in the next section.) 655 SB
1360 2898 46 (In ) 50 SB
1410 2898 105 (many ) 109 SB
1519 2898 145 (existing ) 149 SB
1668 2898 129 (natural ) 134 SB
1802 2898 164 (language ) 169 SB
1971 2898 180 (processing) 180 SB
1210 2948 155 (systems, ) 159 SB
1369 2948 63 (the ) 67 SB
1436 2948 63 (list ) 67 SB
1503 2948 46 (of ) 50 SB
1553 2948 165 (available ) 170 SB
1723 2948 180 (classifiers ) 185 SB
1908 2948 60 (for ) 65 SB
1973 2948 89 (each ) 94 SB
2067 2948 84 (noun) 84 SB
1210 2998 39 (is ) 54 SB
1264 2998 153 (attached ) 168 SB
1432 2998 44 (to ) 59 SB
1491 2998 30 (a ) 45 SB
1536 2998 136 (lexicon ) 151 SB
1687 2998 97 (base. ) 112 SB
1799 2998 107 (Rules ) 123 SB
1922 2998 60 (for ) 76 SB
1998 2998 153 (classifier) 153 SB
1210 3048 162 (selection ) 180 SB
1390 3048 93 (from ) 111 SB
1501 3048 63 (the ) 81 SB
1582 3048 63 (list ) 81 SB
1663 3048 70 (can ) 88 SB
1751 3048 172 (somehow ) 190 SB
1941 3048 140 (provide ) 158 SB
2099 3048 52 (the) 52 SB
1 #C
statusdict begin /manualfeed false store end
EJ RS
%%PageTrailer
%%PageResources: font MSTT31c146
%%+ font Times-Bold
%%+ font Times-Italic
%%+ font Times-Roman
%%Page: 2 2
%%PageResources: (atend)
SS
0 0 25 31 776 1169 300 SM
32 0 0 42 42 0 0 0 38 /Times-Roman /font32 ANSIFont font
0 0 0 fC
165 210 129 (default ) 172 SB
337 210 103 (value ) 147 SB
484 210 65 (but ) 109 SB
593 210 88 (does ) 132 SB
725 210 65 (not ) 109 SB
834 210 176 (guarantee ) 220 SB
1054 210 52 (the) 52 SB
165 260 287 (appropriateness. ) 295 SB
460 260 176 (However, ) 184 SB
644 260 63 (the ) 71 SB
715 260 168 (problems ) 176 SB
891 260 53 (on ) 62 SB
953 260 153 (classifier) 153 SB
165 310 708 (phrase construction still remain unsolved.) 708 SB
315 360 58 (To ) 95 SB
410 360 178 (overcome ) 215 SB
625 360 63 (the ) 100 SB
725 360 168 (problems ) 206 SB
931 360 46 (of ) 84 SB
1015 360 91 (using) 91 SB
165 410 191 (classifiers, ) 199 SB
364 410 60 (we ) 68 SB
432 410 144 (propose ) 152 SB
584 410 30 (a ) 38 SB
622 410 138 (method ) 146 SB
768 410 46 (of ) 55 SB
823 410 164 (classifier ) 173 SB
996 410 110 (phrase) 110 SB
165 460 181 (extracting ) 193 SB
358 460 93 (from ) 105 SB
463 460 30 (a ) 42 SB
505 460 96 (large ) 108 SB
613 460 134 (corpus. ) 146 SB
759 460 57 (As ) 69 SB
828 460 30 (a ) 42 SB
870 460 116 (result, ) 129 SB
999 460 107 (Noun-) 107 SB
165 510 173 (Classifier ) 189 SB
354 510 226 (Associations ) 242 SB
596 510 187 (\(described ) 203 SB
799 510 44 (in ) 61 SB
860 510 138 (Section ) 155 SB
1015 510 46 (3\) ) 63 SB
1078 510 28 (is) 28 SB
165 560 205 (statistically ) 209 SB
374 560 134 (created ) 138 SB
512 560 44 (to ) 48 SB
560 560 117 (define ) 121 SB
681 560 63 (the ) 68 SB
749 560 211 (relationship ) 216 SB
965 560 141 (between) 141 SB
165 610 30 (a ) 33 SB
198 610 95 (noun ) 98 SB
296 610 72 (and ) 75 SB
371 610 30 (a ) 33 SB
404 610 164 (classifier ) 167 SB
571 610 44 (in ) 47 SB
618 610 30 (a ) 33 SB
651 610 164 (classifier ) 167 SB
818 610 132 (phrase. ) 136 SB
954 610 96 (With ) 100 SB
1054 610 52 (the) 52 SB
165 660 180 (frequency ) 199 SB
364 660 46 (of ) 65 SB
429 660 63 (the ) 82 SB
511 660 197 (occurrence ) 216 SB
727 660 46 (of ) 65 SB
792 660 30 (a ) 49 SB
841 660 164 (classifier ) 183 SB
1024 660 44 (in ) 63 SB
1087 660 19 (a) 19 SB
165 710 164 (classifier ) 167 SB
332 710 132 (phrase, ) 135 SB
467 710 60 (we ) 63 SB
530 710 70 (can ) 73 SB
603 710 144 (propose ) 147 SB
750 710 63 (the ) 66 SB
816 710 93 (most ) 97 SB
913 710 193 (appropriate) 193 SB
165 760 67 (use ) 83 SB
248 760 46 (of ) 62 SB
310 760 11 ( ) 27 SB
337 760 30 (a ) 46 SB
383 760 175 (classifier. ) 191 SB
574 760 233 (Furthermore, ) 249 SB
823 760 60 (we ) 76 SB
899 760 171 (introduce ) 188 SB
1087 760 19 (a) 19 SB
165 810 171 (hierarchy ) 184 SB
349 810 46 (of ) 59 SB
408 810 162 (semantic ) 176 SB
584 810 93 (class ) 107 SB
691 810 60 (for ) 74 SB
765 810 63 (the ) 77 SB
842 810 171 (induction ) 185 SB
1027 810 46 (of ) 60 SB
1087 810 19 (a) 19 SB
165 860 164 (classifier ) 172 SB
337 860 93 (class ) 101 SB
438 860 102 (when ) 110 SB
548 860 84 (they ) 92 SB
640 860 63 (are ) 71 SB
711 860 178 (employed ) 187 SB
898 860 44 (to ) 53 SB
951 860 155 (construct) 155 SB
165 910 642 (with nouns which belong to the same ) 643 SB
808 910 93 (class ) 94 SB
902 910 46 (of ) 47 SB
949 910 157 (meaning.) 157 SB
165 960 941 (Section 3 and Section 4 describe the generation and the) 941 SB
165 1010 713 (implementation of the NCA, respectively.) 713 SB
32 0 0 46 46 0 0 0 44 /Times-Bold /font29 ANSIFont font
165 1107 810 (2. The roles of classifier in Thai language) 810 SB
32 0 0 42 42 0 0 0 38 /Times-Roman /font32 ANSIFont font
165 1216 46 (In ) 68 SB
233 1216 89 (Thai ) 111 SB
344 1216 175 (language, ) 197 SB
541 1216 60 (we ) 82 SB
623 1216 67 (use ) 89 SB
712 1216 180 (classifiers ) 203 SB
915 1216 44 (in ) 67 SB
982 1216 124 (various) 124 SB
165 1266 184 (situations. ) 195 SB
360 1266 77 (The ) 88 SB
448 1266 164 (classifier ) 175 SB
623 1266 100 (plays ) 111 SB
734 1266 51 (an ) 62 SB
796 1266 176 (important ) 188 SB
984 1266 77 (role ) 89 SB
1073 1266 33 (in) 33 SB
165 1316 306 (construction with ) 307 SB
472 1316 95 (noun ) 96 SB
568 1316 44 (to ) 45 SB
613 1316 137 (express ) 138 SB
751 1316 142 (ordinal, ) 143 SB
894 1316 162 (pronoun, ) 163 SB
1057 1316 49 (for) 49 SB
165 1366 161 (instance. ) 193 SB
358 1366 77 (The ) 110 SB
468 1366 164 (classifier ) 197 SB
665 1366 121 (phrase ) 154 SB
819 1366 39 (is ) 72 SB
891 1366 215 (syntactically) 215 SB
165 1416 176 (generated ) 189 SB
354 1416 178 (according ) 191 SB
545 1416 44 (to ) 58 SB
603 1416 30 (a ) 44 SB
647 1416 143 (specific ) 157 SB
804 1416 140 (pattern. ) 154 SB
32 0 0 42 42 0 0 0 41 /Times-Bold /font29 ANSIFont font
958 1413 81 (Fig. ) 95 SB
1053 1413 53 (2.1) 53 SB
32 0 0 42 42 0 0 0 38 /Times-Roman /font32 ANSIFont font
165 1467 941 (shows the position of a classifier in each pattern, where) 941 SB
165 1517 41 (N ) 43 SB
208 1517 116 (stands ) 118 SB
326 1517 60 (for ) 62 SB
388 1517 106 (noun, ) 108 SB
496 1517 136 (NCNM ) 138 SB
634 1517 116 (stands ) 118 SB
752 1517 60 (for ) 63 SB
815 1517 148 (cardinal ) 151 SB
966 1517 140 (number,) 140 SB
165 1567 65 (CL ) 73 SB
238 1567 116 (stands ) 124 SB
362 1567 60 (for ) 69 SB
431 1567 175 (classifier, ) 184 SB
615 1567 93 (DET ) 102 SB
717 1567 116 (stands ) 125 SB
842 1567 60 (for ) 69 SB
911 1567 195 (determiner,) 195 SB
165 1617 123 (VATT ) 129 SB
294 1617 116 (stands ) 122 SB
416 1617 60 (for ) 66 SB
482 1617 186 (attributive ) 192 SB
674 1617 97 (verb, ) 104 SB
778 1617 149 (REL_M ) 156 SB
934 1617 116 (stands ) 123 SB
1057 1617 49 (for) 49 SB
165 1667 139 (relative ) 158 SB
323 1667 142 (marker, ) 161 SB
484 1667 137 (ITR_M ) 156 SB
640 1667 11 ( ) 31 SB
671 1667 116 (stands ) 136 SB
807 1667 60 (for ) 80 SB
887 1667 219 (Interrogative) 219 SB
165 1717 131 (marker ) 137 SB
302 1717 22 (, ) 28 SB
330 1717 138 (DONM ) 144 SB
474 1717 11 ( ) 18 SB
492 1717 116 (stands ) 123 SB
615 1717 60 (for ) 67 SB
682 1717 131 (ordinal ) 138 SB
820 1717 161 (numeral, ) 168 SB
988 1717 118 (DDAC) 118 SB
165 1767 557 (stands for definite demonstrative) 557 SB
315 1817 109 (Study ) 132 SB
447 1817 53 (on ) 77 SB
524 1817 63 (the ) 87 SB
611 1817 67 (use ) 91 SB
702 1817 46 (of ) 70 SB
772 1817 164 (classifier ) 188 SB
960 1817 44 (in ) 68 SB
1028 1817 78 (each) 78 SB
165 1867 191 (expression ) 193 SB
358 1867 190 (mentioned ) 192 SB
550 1867 123 (above, ) 125 SB
675 1867 60 (we ) 62 SB
737 1867 70 (can ) 72 SB
809 1867 164 (conclude ) 167 SB
976 1867 75 (that ) 78 SB
1054 1867 52 (the) 52 SB
165 1917 100 (types ) 109 SB
274 1917 46 (of ) 55 SB
329 1917 164 (classifier ) 173 SB
502 1917 63 (are ) 72 SB
574 1917 65 (not ) 74 SB
648 1917 169 (restricted ) 178 SB
826 1917 44 (to ) 53 SB
879 1917 72 (any ) 81 SB
960 1917 102 (kinds ) 111 SB
1071 1917 35 (of) 35 SB
165 1967 202 (expression. ) 204 SB
369 1967 58 (To ) 60 SB
429 1967 154 (consider ) 157 SB
586 1967 63 (the ) 66 SB
652 1967 162 (semantic ) 165 SB
817 1967 251 (representation ) 254 SB
1071 1967 35 (of) 35 SB
165 2017 89 (each ) 96 SB
261 2017 202 (expression, ) 209 SB
470 2017 35 (it ) 42 SB
512 2017 149 (happens ) 156 SB
668 2017 75 (that ) 82 SB
750 2017 63 (the ) 71 SB
821 2017 77 (unit ) 85 SB
906 2017 164 (classifier ) 172 SB
1078 2017 28 (is) 28 SB
165 2067 65 (not ) 74 SB
239 2067 159 (regarded ) 168 SB
407 2067 46 (as ) 55 SB
462 2067 30 (a ) 39 SB
501 2067 195 (conceptual ) 204 SB
705 2067 77 (unit ) 87 SB
792 2067 44 (in ) 54 SB
846 2067 54 (all ) 64 SB
910 2067 196 (expressions) 196 SB
165 2117 122 (except ) 137 SB
302 2117 44 (in ) 59 SB
361 2117 129 (pattern ) 144 SB
505 2117 43 (6, ) 58 SB
563 2117 65 (but ) 80 SB
643 2117 63 (the ) 78 SB
721 2117 98 (other ) 113 SB
834 2117 100 (types ) 115 SB
949 2117 74 (are. ) 89 SB
1038 2117 68 (\(see) 68 SB
165 2167 374 (examples in a. and b.\)) 374 SB
315 2273 44 (a\) ) 44 SB
32 0 0 50 50 0 0 1 47 /MSTT31c146 font

/Gbb [22.0 0.0 1.0 0.0 20.0 32.0]
/Gbb {
    19 32 true [1 0 0 -1 -1.0 32.0] {<0000600000e00000e00000e00000e00000e00000e00000e00000e00000e00000e00000e03c00e07e
00e0c700e08700e04f00e07f00e00700e00700e00700e00700e00700e00700e00700e00700e00700
e00700e00700e00700e00701e01fff80>} imagemask 
  }
  187 /Gbb MSTT31c146 AddChar
/Gaa [20.0 0.0 1.0 0.0 20.0 20.0]
/Gaa {
    19 20 true [1 0 0 -1 -1.0 20.0] {<0f00e03fc0e061c1c040e1c048e300dcfe0046fe0064e7003ce30000e38000e38000e38000e38000
e38000e38000e38000e30000e70000e60007fc00>} imagemask 
  }
  170 /Gaa MSTT31c146 AddChar
/Ga4 [20.0 0.0 2.0 0.0 17.0 20.0]
/Ga4 {
    15 20 true [1 0 0 -1 -2.0 20.0] {<07e01df8203c401ec40e8f0e988e988ef88e7d8e3b0e380e380e380e380e380e380e380e380e380e
>} imagemask 
  }
  164 /Ga4 MSTT31c146 AddChar

359 2264 334 ( \273\303\320\252\322\252\271     \312\315\247   \244\271) 334 SB
32 0 0 42 42 0 0 0 38 /Times-Roman /font32 ANSIFont font
315 2334 423 (     /prachachon  2  khon/) 423 SB
465 2384 303 (            \(Unit-CL\)) 303 SB
315 2434 482 (         people      2  <people>) 482 SB
315 2484 227 (     '2  people') 227 SB
315 2590 57 (b\)  ) 57 SB
32 0 0 50 50 0 0 1 47 /MSTT31c146 font

/Gd8 [0.0 0.0 -10.0 -9.0 -3.0 -1.0]
/Gd8 {
    7 8 true [1 0 0 -1 10.0 -1.0] {<7c4e8e8e7e0e0e0e>} imagemask 
  }
  216 /Gd8 MSTT31c146 AddChar
/G8b [0.0 0.0 -6.0 25.0 -2.0 32.0]
/G8b {
    4 7 true [1 0 0 -1 6.0 32.0] {<f0f0f0f0f06060>} imagemask 
  }
  139 /G8b MSTT31c146 AddChar

372 2581 331 (\273\303\320\252\322\252\271     \312\315\247   \241\305\330\213 ) 320 SB
692 2581 21 (\301) 21 SB
32 0 0 42 42 0 0 0 38 /Times-Roman /font32 ANSIFont font
315 2651 426 (     /prachachon  2  klum/) 426 SB
465 2701 403 (            \(Collective-CL\)) 403 SB
315 2751 467 (         people      2  <group>) 467 SB
315 2801 398 (     '2  groups of people') 398 SB
165 2901 11 ( ) 11 SB
315 2901 70 (We ) 79 SB
394 2901 218 (encountered ) 227 SB
621 2901 44 (to ) 54 SB
675 2901 155 (generate ) 165 SB
840 2901 63 (the ) 73 SB
913 2901 193 (appropriate) 193 SB
165 2951 495 (classifier for noun or verb in ) 496 SB
661 2951 30 (a ) 31 SB
692 2951 162 (semantic ) 163 SB
855 2951 251 (representation.) 251 SB
165 3001 77 (The ) 121 SB
286 3001 164 (classifier ) 208 SB
494 3001 201 (assignment ) 246 SB
740 3001 60 (for ) 105 SB
845 3001 261 (non-conceptual) 261 SB
165 3051 251 (representation ) 262 SB
427 3051 72 (and ) 83 SB
510 3051 63 (the ) 74 SB
584 3051 164 (classifier ) 175 SB
759 3051 162 (selection ) 173 SB
932 3051 46 (of ) 57 SB
989 3051 72 (one ) 84 SB
1073 3051 33 (to) 33 SB
1210 210 551 (many conceptual representation ) 552 SB
1762 210 63 (are ) 64 SB
1826 210 86 (over ) 87 SB
1913 210 195 (handleable ) 196 SB
2109 210 42 (by) 42 SB
1210 260 63 (the ) 77 SB
1287 260 187 (rule-based ) 201 SB
1488 260 177 (approach. ) 191 SB
1679 260 77 (The ) 92 SB
1771 260 144 (propose ) 159 SB
1930 260 53 (on ) 68 SB
1998 260 153 (classifier) 153 SB
1210 310 201 (assignment ) 204 SB
1414 310 102 (using ) 106 SB
1520 310 63 (the ) 67 SB
1587 310 233 (corpus-based ) 237 SB
1824 310 138 (method ) 142 SB
1966 310 11 ( ) 15 SB
1981 310 39 (is ) 43 SB
2024 310 127 (another) 127 SB
1210 360 177 (approach. ) 195 SB
1405 360 114 (Based ) 133 SB
1538 360 53 (on ) 72 SB
1610 360 63 (the ) 82 SB
1692 360 200 (collocation ) 219 SB
1911 360 46 (of ) 65 SB
1976 360 95 (noun ) 114 SB
2090 360 61 (and) 61 SB
1210 410 592 (classifier of each pattern shown in ) 592 SB
32 0 0 42 42 0 0 0 41 /Times-Bold /font29 ANSIFont font
1802 407 134 (Fig. 2.1) 134 SB
32 0 0 42 42 0 0 0 38 /Times-Roman /font32 ANSIFont font
1936 410 82 (, we ) 83 SB
2019 410 132 (decided) 132 SB
1210 461 44 (to ) 46 SB
1256 461 166 (construct ) 168 SB
1424 461 63 (the ) 66 SB
1490 461 104 (Noun ) 107 SB
1597 461 173 (Classifier ) 176 SB
1773 461 210 (Association ) 213 SB
1986 461 94 (table ) 97 SB
2083 461 68 (\(see) 68 SB
1210 511 138 (Section ) 148 SB
1358 511 57 (3\). ) 67 SB
1425 511 41 (A ) 51 SB
1476 511 178 (stochastic ) 188 SB
1664 511 138 (method ) 149 SB
1813 511 178 (combined ) 189 SB
2002 511 86 (with ) 97 SB
2099 511 52 (the) 52 SB
1210 561 143 (concept ) 145 SB
1355 561 171 (hierarchy ) 173 SB
1528 561 39 (is ) 41 SB
1569 561 165 (proposed ) 167 SB
1736 561 11 ( ) 13 SB
1749 561 46 (as ) 48 SB
1797 561 30 (a ) 32 SB
1829 561 145 (strategy ) 148 SB
1977 561 44 (in ) 47 SB
2024 561 127 (making) 127 SB
1210 611 344 (the NCA table. The ) 345 SB
1555 611 94 (table ) 95 SB
1650 611 177 (composes ) 178 SB
1828 611 46 (of ) 47 SB
1875 611 11 ( ) 12 SB
1887 611 63 (the ) 64 SB
1951 611 200 (information) 200 SB
1210 661 105 (about ) 110 SB
1320 661 262 (noun-classifier ) 267 SB
1587 661 211 (collocation, ) 216 SB
1803 661 141 (statistic ) 146 SB
1949 661 202 (occurrences) 202 SB
1210 711 941 (and the representative classifier for each semantic class) 941 SB
1210 761 421 (in the concept hierarchy.) 421 SB
32 0 0 46 46 0 0 0 44 /Times-Bold /font29 ANSIFont font
1210 858 864 (3. Extraction of Noun-Classifier Collocation) 864 SB
32 0 0 42 42 0 0 0 38 /Times-Roman /font32 ANSIFont font
1210 967 46 (In ) 57 SB
1267 967 72 (this ) 83 SB
1350 967 142 (section, ) 153 SB
1503 967 60 (we ) 72 SB
1575 967 152 (describe ) 164 SB
1739 967 63 (the ) 75 SB
1814 967 176 (algorithm ) 188 SB
2002 967 88 (used ) 100 SB
2102 967 49 (for) 49 SB
1210 1017 730 (extraction of Noun Classifier Associations ) 731 SB
1941 1017 127 (\(NCA\) ) 128 SB
2069 1017 82 (from) 82 SB
1210 1067 30 (a ) 37 SB
1247 1067 96 (large ) 103 SB
1350 1067 134 (corpus. ) 141 SB
1491 1067 70 (We ) 77 SB
1568 1067 88 (used ) 95 SB
1663 1067 30 (a ) 37 SB
1700 1067 53 (40 ) 60 SB
1760 1067 176 (megabyte ) 183 SB
1943 1067 89 (Thai ) 96 SB
2039 1067 112 (corpus) 112 SB
1210 1117 165 (collected ) 172 SB
1382 1117 93 (from ) 100 SB
1482 1117 135 (various ) 142 SB
1624 1117 98 (areas ) 105 SB
1729 1117 44 (to ) 51 SB
1780 1117 113 (create ) 121 SB
1901 1117 63 (the ) 71 SB
1972 1117 105 (table. ) 113 SB
2085 1117 66 (The) 66 SB
1210 1167 399 (algorithm is as follows:) 399 SB
32 0 0 42 42 0 0 0 41 /Times-Bold /font29 ANSIFont font
1210 1264 111 (Step 1) 111 SB
32 0 0 42 42 0 0 0 38 /Times-Roman /font32 ANSIFont font
1321 1267 12 (:) 12 SB
1360 1267 344 (Word segmentation.) 344 SB
1210 1318 101 (Input:) 101 SB
1360 1318 164 (A corpus.) 164 SB
1210 1368 129 (Output:) 129 SB
1360 1368 492 (The word-segmented corpus.) 492 SB
1210 1468 46 (In ) 62 SB
1272 1468 75 (text ) 91 SB
1363 1468 202 (processing, ) 218 SB
1581 1468 60 (we ) 76 SB
1657 1468 98 (often ) 114 SB
1771 1468 91 (need ) 107 SB
1878 1468 97 (word ) 114 SB
1992 1468 159 (boundary) 159 SB
1210 1518 211 (information ) 213 SB
1423 1518 60 (for ) 62 SB
1485 1518 131 (several ) 133 SB
1618 1518 171 (purposes. ) 174 SB
1792 1518 152 (Because ) 155 SB
1947 1518 89 (Thai ) 92 SB
2039 1518 67 (has ) 70 SB
2109 1518 42 (no) 42 SB
1210 1568 891 (explicit marker to separate words from one another, ) 892 SB
2102 1568 49 (we) 49 SB
1210 1618 91 (have ) 92 SB
1302 1618 44 (to ) 45 SB
1347 1618 191 (preprocess ) 192 SB
1539 1618 63 (the ) 64 SB
1603 1618 123 (corpus ) 124 SB
1727 1618 86 (with ) 87 SB
1814 1618 11 ( ) 12 SB
1826 1618 97 (word ) 99 SB
1925 1618 226 (segmentation) 226 SB
1210 1668 165 (program. ) 194 SB
1404 1668 70 (We ) 99 SB
1503 1668 88 (used ) 117 SB
1620 1668 63 (the ) 92 SB
1712 1668 154 (program ) 183 SB
1895 1668 185 (developed ) 214 SB
2109 1668 42 (by) 42 SB
1210 1718 324 (Sornlertlamvanich ) 330 SB
1540 1718 123 (\(1993\) ) 129 SB
1669 1718 86 (with ) 93 SB
1762 1718 213 (post-editing ) 220 SB
1982 1718 44 (to ) 51 SB
2033 1718 118 (correct) 118 SB
1210 1768 89 (fault ) 99 SB
1309 1768 248 (segmentation. ) 258 SB
1567 1768 11 ( ) 21 SB
1588 1768 77 (The ) 87 SB
1675 1768 154 (program ) 165 SB
1840 1768 154 (employs ) 165 SB
2005 1768 146 (heuristic) 146 SB
1210 1818 93 (rules ) 114 SB
1324 1818 46 (of ) 67 SB
1391 1818 133 (longest ) 154 SB
1545 1818 169 (matching ) 190 SB
1735 1818 72 (and ) 93 SB
1828 1818 89 (least ) 110 SB
1938 1818 97 (word ) 119 SB
2057 1818 94 (count) 94 SB
1210 1868 225 (incorporated ) 231 SB
1441 1868 86 (with ) 92 SB
1533 1868 167 (character ) 174 SB
1707 1868 192 (combining ) 199 SB
1906 1868 93 (rules ) 100 SB
2006 1868 60 (for ) 67 SB
2073 1868 78 (Thai) 78 SB
1210 1918 135 (words.  ) 136 SB
1346 1918 142 (Though ) 143 SB
1489 1918 63 (the ) 64 SB
1553 1918 162 (accuracy ) 163 SB
1716 1918 46 (of ) 47 SB
1763 1918 63 (the ) 64 SB
1827 1918 97 (word ) 98 SB
1925 1918 226 (segmentation) 226 SB
1210 1968 256 (does not reach ) 257 SB
1467 1968 120 (100%, ) 121 SB
1588 1968 65 (but ) 66 SB
1654 1968 35 (it ) 36 SB
1690 1968 39 (is ) 40 SB
1730 1968 86 (high ) 87 SB
1817 1968 135 (enough ) 136 SB
1953 1968 11 ( ) 12 SB
1965 1968 112 (\(more ) 113 SB
2078 1968 73 (than) 73 SB
1210 2018 633 (95%\) to reduce the post-editing time.) 633 SB
32 0 0 42 42 0 0 0 41 /Times-Bold /font29 ANSIFont font
1210 2115 111 (Step 2) 111 SB
32 0 0 42 42 0 0 0 38 /Times-Roman /font32 ANSIFont font
1321 2118 12 (:) 12 SB
1360 2118 152 (Tagging.) 152 SB
1210 2169 101 (Input:) 101 SB
1360 2169 285 (Output of step 1.) 285 SB
1210 2219 129 (Output:) 129 SB
1360 2219 784 (The corpus of which each word is tagged with) 784 SB
1210 2269 636 (a part of speech and a semantic class.) 636 SB
1210 2369 77 (The ) 83 SB
1293 2369 292 (word-segmented ) 298 SB
1591 2369 123 (corpus ) 129 SB
1720 2369 39 (is ) 45 SB
1765 2369 84 (then ) 90 SB
1855 2369 177 (processed ) 184 SB
2039 2369 86 (with ) 93 SB
2132 2369 19 (a) 19 SB
1210 2419 178 (stochastic ) 179 SB
1389 2419 255 (part-of-speech ) 256 SB
1645 2419 128 (tagger. ) 129 SB
1774 2419 96 (Each ) 98 SB
1872 2419 97 (word ) 99 SB
32 0 0 42 42 0 0 0 39 /Times-Italic /font31 ANSIFont font
1971 2418 28 (w) 28 SB
32 0 0 42 42 0 0 0 38 /Times-Roman /font32 ANSIFont font
1999 2419 11 ( ) 13 SB
2012 2419 139 (together) 139 SB
1210 2471 86 (with ) 96 SB
1306 2471 51 (its ) 61 SB
1367 2471 77 (part ) 87 SB
1454 2471 46 (of ) 57 SB
1511 2471 126 (speech ) 137 SB
1648 2471 39 (is ) 50 SB
1698 2471 84 (then ) 95 SB
1793 2471 88 (used ) 99 SB
1892 2471 44 (to ) 55 SB
1947 2471 141 (retrieve ) 152 SB
2099 2471 52 (the) 52 SB
1210 2521 162 (semantic ) 174 SB
1384 2521 93 (class ) 105 SB
1489 2521 46 (of ) 58 SB
1547 2521 63 (the ) 75 SB
1622 2521 97 (word ) 109 SB
1731 2521 93 (from ) 105 SB
1836 2521 30 (a ) 42 SB
1878 2521 194 (dictionary. ) 207 SB
2085 2521 66 (The) 66 SB
1210 2571 105 (result ) 121 SB
1331 2571 112 (yields ) 129 SB
1460 2571 30 (a ) 47 SB
1507 2571 82 (data ) 99 SB
1606 2571 159 (structure ) 176 SB
1782 2571 46 (of ) 63 SB
32 0 0 42 42 0 0 0 39 /Times-Italic /font31 ANSIFont font
1845 2570 115 (\(w,p,s\)) 115 SB
32 0 0 42 42 0 0 0 38 /Times-Roman /font32 ANSIFont font
1960 2571 22 (, ) 39 SB
1999 2571 114 (where ) 131 SB
32 0 0 42 42 0 0 0 39 /Times-Italic /font31 ANSIFont font
2130 2570 21 (p) 21 SB
32 0 0 42 42 0 0 0 38 /Times-Roman /font32 ANSIFont font
1210 2623 140 (denotes ) 151 SB
1361 2623 63 (the ) 74 SB
1435 2623 77 (part ) 88 SB
1523 2623 46 (of ) 57 SB
1580 2623 126 (speech ) 137 SB
1717 2623 46 (of ) 57 SB
32 0 0 42 42 0 0 0 39 /Times-Italic /font31 ANSIFont font
1774 2622 28 (w) 28 SB
32 0 0 42 42 0 0 0 38 /Times-Roman /font32 ANSIFont font
1802 2623 11 ( ) 22 SB
1824 2623 72 (and ) 84 SB
32 0 0 42 42 0 0 0 39 /Times-Italic /font31 ANSIFont font
1908 2622 16 (s) 16 SB
32 0 0 42 42 0 0 0 38 /Times-Roman /font32 ANSIFont font
1924 2623 11 ( ) 23 SB
1947 2623 140 (denotes ) 152 SB
2099 2623 52 (the) 52 SB
1210 2675 162 (semantic ) 163 SB
1373 2675 93 (class ) 94 SB
1467 2675 46 (of ) 48 SB
32 0 0 42 42 0 0 0 39 /Times-Italic /font31 ANSIFont font
1515 2674 28 (w) 28 SB
32 0 0 42 42 0 0 0 38 /Times-Roman /font32 ANSIFont font
1543 2675 22 (. ) 24 SB
1567 2675 69 (For ) 71 SB
1638 2675 166 (example, ) 168 SB
1806 2675 63 (the ) 65 SB
1871 2675 82 (data ) 84 SB
1955 2675 159 (structure ) 161 SB
2116 2675 35 (of) 35 SB
1210 2733 63 (the ) 67 SB
1277 2733 97 (word ) 101 SB
32 0 0 50 50 0 0 1 47 /MSTT31c146 font

/Gd1 [0.0 0.0 -12.0 24.0 2.0 31.0]
/Gd1 {
    14 7 true [1 0 0 -1 12.0 31.0] {<3c0c660cc208c218c6307de03fc0>} imagemask 
  }
  209 /Gd1 MSTT31c146 AddChar
/Gc2 [17.0 0.0 1.0 0.0 15.0 20.0]
/Gc2 {
    14 20 true [1 0 0 -1 -1.0 20.0] {<1c0c361c631ce31cf21cde1c401c601c3e1c701ce01ce01ce01ce01ce01ce01c601c70383ef00fc0
>} imagemask 
  }
  194 /Gc2 MSTT31c146 AddChar

1378 2724 34 (\271\321 ) 23 SB
1401 2724 56 (\241\340\303\325 ) 45 SB
1446 2724 51 (\302\271 ) 55 SB
32 0 0 42 42 0 0 0 38 /Times-Roman /font32 ANSIFont font
1501 2733 162 (/nakrian/ ) 167 SB
1668 2733 149 ('student' ) 154 SB
1822 2733 39 (is ) 44 SB
1866 2733 14 (\() 14 SB
32 0 0 50 50 0 0 1 47 /MSTT31c146 font
1880 2724 34 (\271\321 ) 23 SB
1903 2724 56 (\241\340\303\325 ) 45 SB
1948 2724 40 (\302\271) 40 SB
32 0 0 42 42 0 0 0 38 /Times-Roman /font32 ANSIFont font
1988 2733 22 (, ) 27 SB
2015 2733 136 (NCMN,) 136 SB
1210 2794 148 (person\), ) 155 SB
1365 2794 114 (where ) 121 SB
1486 2794 136 (NCMN ) 143 SB
1629 2794 116 (stands ) 123 SB
1752 2794 60 (for ) 68 SB
1820 2794 159 (common ) 167 SB
1987 2794 95 (noun ) 103 SB
2090 2794 61 (and) 61 SB
1210 2850 305 (person represents ) 305 SB
32 0 0 50 50 0 0 1 47 /MSTT31c146 font
1515 2841 34 (\271\321 ) 23 SB
1538 2841 56 (\241\340\303\325 ) 45 SB
1583 2841 51 (\302\271 ) 51 SB
32 0 0 42 42 0 0 0 38 /Times-Roman /font32 ANSIFont font
1634 2850 380 ( in the class of person.) 380 SB
32 0 0 42 42 0 0 0 41 /Times-Bold /font29 ANSIFont font
1210 2958 111 (Step 3) 111 SB
32 0 0 42 42 0 0 0 38 /Times-Roman /font32 ANSIFont font
1321 2961 12 (:) 12 SB
1360 2961 425 (Producing concordances.) 425 SB
1210 3012 101 (Input:) 101 SB
1360 3012 595 (Output of step 2, a given classifier ) 595 SB
32 0 0 42 42 0 0 0 39 /Times-Italic /font31 ANSIFont font
1955 3011 42 (cl.) 42 SB
32 0 0 42 42 0 0 0 38 /Times-Roman /font32 ANSIFont font
1210 3064 129 (Output:) 129 SB
1360 3064 498 (All the fragments containing ) 498 SB
32 0 0 42 42 0 0 0 39 /Times-Italic /font31 ANSIFont font
1858 3063 42 (cl.) 42 SB
1 #C
statusdict begin /manualfeed false store end
EJ RS
%%PageTrailer
%%PageResources: font MSTT31c146
%%+ font Times-Bold
%%+ font Times-Italic
%%+ font Times-Roman
%%Page: 3 3
%%PageResources: (atend)
SS
0 0 25 31 776 1169 300 SM
0 0 0 fC
/fm 256 def
2 2 390 299 B
1 F
n
/fm 256 def
2 2 390 299 B
1 F
n
/fm 256 def
414 2 393 299 B
1 F
n
/fm 256 def
2 2 808 299 B
1 F
n
/fm 256 def
353 2 811 299 B
1 F
n
/fm 256 def
2 2 1165 299 B
1 F
n
/fm 256 def
825 2 1168 299 B
1 F
n
/fm 256 def
2 2 1994 299 B
1 F
n
/fm 256 def
2 2 1994 299 B
1 F
n
/fm 256 def
2 45 390 302 B
1 F
n
/fm 256 def
2 45 808 302 B
1 F
n
/fm 256 def
2 45 1165 302 B
1 F
n
/fm 256 def
2 45 1994 302 B
1 F
n
32 0 0 38 38 0 0 0 37 /Times-Bold /font29 ANSIFont font
gs 415 49 393 299 CB
502 302 195 (Expressions) 195 SB
gr
gs 354 49 811 299 CB
917 302 138 (Patterns) 138 SB
gr
gs 826 49 1168 299 CB
1516 302 136 (Samples) 136 SB
gr
/fm 256 def
2 2 390 348 B
1 F
n
/fm 256 def
414 2 393 348 B
1 F
n
/fm 256 def
2 2 808 348 B
1 F
n
/fm 256 def
353 2 811 348 B
1 F
n
/fm 256 def
2 2 1165 348 B
1 F
n
/fm 256 def
825 2 1168 348 B
1 F
n
/fm 256 def
2 2 1994 348 B
1 F
n
/fm 256 def
2 183 390 351 B
1 F
n
/fm 256 def
2 183 808 351 B
1 F
n
/fm 256 def
2 183 1165 351 B
1 F
n
/fm 256 def
2 183 1994 351 B
1 F
n
32 0 0 38 38 0 0 0 34 /Times-Roman /font32 ANSIFont font
gs 415 49 393 348 CB
414 354 237 (1. Enumeration) 237 SB
gr
gs 354 49 811 348 CB
833 354 252 (N/V-NCNM-CL) 252 SB
gr
gs 826 49 1168 348 CB
1190 354 322 ( /nakrian    3    khon/) 322 SB
gr
gs 826 49 1168 394 CB
1190 400 340 (      \(N\)       \(N\)   \(CL\)) 340 SB
gr
gs 826 49 1168 440 CB
1190 445 10 ( ) 10 SB
gr
32 0 0 38 38 0 0 0 35 /Times-Italic /font31 ANSIFont font
gs 826 49 1168 440 CB
1200 444 363 (student    3   <student>) 363 SB
gr
32 0 0 38 38 0 0 0 34 /Times-Roman /font32 ANSIFont font
gs 826 49 1168 486 CB
1190 492 237 ( 'three students') 237 SB
gr
/fm 256 def
2 2 390 535 B
1 F
n
/fm 256 def
414 2 393 535 B
1 F
n
/fm 256 def
2 2 808 535 B
1 F
n
/fm 256 def
353 2 811 535 B
1 F
n
/fm 256 def
2 2 1165 535 B
1 F
n
/fm 256 def
825 2 1168 535 B
1 F
n
/fm 256 def
2 2 1994 535 B
1 F
n
/fm 256 def
2 183 390 538 B
1 F
n
/fm 256 def
2 183 808 538 B
1 F
n
/fm 256 def
2 183 1165 538 B
1 F
n
/fm 256 def
2 183 1994 538 B
1 F
n
gs 415 49 393 535 CB
414 541 156 (2. Ordinal) 156 SB
gr
gs 354 49 811 535 CB
833 541 282 (N-CL-/tii/-NCNM) 282 SB
gr
gs 826 49 1168 535 CB
1190 541 320 (/kaew   bai       thii4/) 320 SB
gr
gs 826 49 1168 581 CB
1190 587 330 (   \(N\)      \(CL\)      \(N\)) 330 SB
gr
32 0 0 38 38 0 0 0 35 /Times-Italic /font31 ANSIFont font
gs 826 49 1168 627 CB
1190 631 329 ( glass   <glass>   4th) 329 SB
gr
32 0 0 38 38 0 0 0 34 /Times-Roman /font32 ANSIFont font
gs 826 49 1168 673 CB
1190 679 262 ( 'the fourth glass') 262 SB
gr
/fm 256 def
2 2 390 722 B
1 F
n
/fm 256 def
414 2 393 722 B
1 F
n
/fm 256 def
2 2 808 722 B
1 F
n
/fm 256 def
353 2 811 722 B
1 F
n
/fm 256 def
2 2 1165 722 B
1 F
n
/fm 256 def
825 2 1168 722 B
1 F
n
/fm 256 def
2 2 1994 722 B
1 F
n
/fm 256 def
2 1011 390 725 B
1 F
n
/fm 256 def
2 1011 808 725 B
1 F
n
/fm 256 def
2 1011 1165 725 B
1 F
n
/fm 256 def
2 1011 1994 725 B
1 F
n
gs 415 49 393 722 CB
414 728 261 (3. Determination) 261 SB
gr
gs 415 49 393 768 CB
414 774 179 (    -Definite) 179 SB
gr
gs 415 49 393 814 CB
414 820 270 (     demonstration) 270 SB
gr
gs 415 49 393 906 CB
414 912 203 (    -Indefinite) 203 SB
gr
gs 415 49 393 952 CB
414 958 270 (     demonstration) 270 SB
gr
gs 415 49 393 1412 CB
414 1418 224 (    -Referential) 224 SB
gr
gs 354 49 811 722 CB
833 728 224 (a\)  N-CL-DET) 224 SB
gr
gs 354 49 811 906 CB
833 912 224 (a\)  N-CL-DET) 224 SB
gr
gs 354 49 811 1228 CB
833 1234 226 (b\)  N-DET-CL) 226 SB
gr
gs 354 49 811 1412 CB
833 1418 224 (a\)  N-CL-DET) 224 SB
gr
gs 826 49 1168 722 CB
1190 728 601 (a\) /raw chop kruangkhidlek kruang nii/) 601 SB
gr
gs 826 49 1168 768 CB
1190 774 726 (                                \(N\)              \(CL\)    \(DET\)) 726 SB
gr
gs 826 49 1168 814 CB
1190 819 50 (     ) 50 SB
gr
32 0 0 38 38 0 0 0 35 /Times-Italic /font31 ANSIFont font
gs 826 49 1168 814 CB
1240 818 572 (we like  calculator <calculator> this) 572 SB
gr
32 0 0 38 38 0 0 0 34 /Times-Roman /font32 ANSIFont font
gs 826 49 1168 860 CB
1190 866 394 (    'we like this calculator') 394 SB
gr
gs 826 49 1168 906 CB
1190 912 585 (a\) /phukhawfung  khon  nung sadaeng) 585 SB
gr
gs 826 49 1168 952 CB
1190 958 516 (            \(N\)               \(CL\)  \(DET\)) 516 SB
gr
32 0 0 38 38 0 0 0 35 /Times-Italic /font31 ANSIFont font
gs 826 49 1168 998 CB
1190 1002 654 (     participant <participant> one  express) 654 SB
gr
32 0 0 38 38 0 0 0 34 /Times-Roman /font32 ANSIFont font
gs 826 49 1168 1044 CB
1190 1050 501 (     khwamhen  nai  thiiprachum/) 501 SB
gr
gs 826 49 1168 1090 CB
1190 1095 50 (     ) 50 SB
gr
32 0 0 38 38 0 0 0 35 /Times-Italic /font31 ANSIFont font
gs 826 49 1168 1090 CB
1240 1094 455 (opinion         in     conference) 455 SB
gr
32 0 0 38 38 0 0 0 34 /Times-Roman /font32 ANSIFont font
gs 826 49 1168 1136 CB
1190 1142 632 (    'A participant expressed his opinion in) 632 SB
gr
gs 826 49 1168 1182 CB
1190 1188 292 (     the conference.') 292 SB
gr
gs 826 49 1168 1228 CB
1190 1234 314 (b\) /sunak  bang  tua/) 314 SB
gr
gs 826 49 1168 1274 CB
1190 1280 396 (        \(N\)      \(DET\)   \(CL\)) 396 SB
gr
gs 826 49 1168 1320 CB
1190 1325 60 (      ) 60 SB
gr
32 0 0 38 38 0 0 0 35 /Times-Italic /font31 ANSIFont font
gs 826 49 1168 1320 CB
1250 1324 334 (dog     some    <dog>) 334 SB
gr
32 0 0 38 38 0 0 0 34 /Times-Roman /font32 ANSIFont font
gs 826 49 1168 1366 CB
1190 1372 217 (    'some dogs') 217 SB
gr
gs 826 49 1168 1412 CB
1190 1418 546 (a\) /kamakan  kana   nii     thukkhon) 546 SB
gr
gs 826 49 1168 1458 CB
1190 1464 456 (           \(N\)         \(CL\)   \(DET\)) 456 SB
gr
32 0 0 38 38 0 0 0 35 /Times-Italic /font31 ANSIFont font
gs 826 49 1168 1504 CB
1190 1508 604 (      committee <group>  this  everyone) 604 SB
gr
32 0 0 38 38 0 0 0 34 /Times-Roman /font32 ANSIFont font
gs 826 49 1168 1550 CB
1190 1556 625 (      chuua  waa    ja     thamngan samret/) 625 SB
gr
32 0 0 38 38 0 0 0 35 /Times-Italic /font31 ANSIFont font
gs 826 49 1168 1596 CB
1190 1600 638 (      believe  that   will    work       success) 638 SB
gr
32 0 0 38 38 0 0 0 34 /Times-Roman /font32 ANSIFont font
gs 826 49 1168 1642 CB
1190 1648 564 (    'It is this committee that everyone) 564 SB
gr
gs 826 49 1168 1688 CB
1190 1694 650 (     believed its mission would be success.') 650 SB
gr
/fm 256 def
2 2 390 1737 B
1 F
n
/fm 256 def
414 2 393 1737 B
1 F
n
/fm 256 def
2 2 808 1737 B
1 F
n
/fm 256 def
353 2 811 1737 B
1 F
n
/fm 256 def
2 2 1165 1737 B
1 F
n
/fm 256 def
825 2 1168 1737 B
1 F
n
/fm 256 def
2 2 1994 1737 B
1 F
n
/fm 256 def
2 183 390 1740 B
1 F
n
/fm 256 def
2 183 808 1740 B
1 F
n
/fm 256 def
2 183 1165 1740 B
1 F
n
/fm 256 def
2 183 1994 1740 B
1 F
n
gs 415 49 393 1737 CB
414 1743 218 (4.  Attributive) 218 SB
gr
gs 354 49 811 1737 CB
833 1743 201 (N-CL-VATT) 201 SB
gr
gs 826 49 1168 1737 CB
1190 1743 320 (/dinsoo   theng   san/) 320 SB
gr
gs 826 49 1168 1783 CB
1190 1789 433 (     \(N\)        \(CL\)     \(VATT\)) 433 SB
gr
32 0 0 38 38 0 0 0 35 /Times-Italic /font31 ANSIFont font
gs 826 49 1168 1829 CB
1190 1833 404 (  pencil   <shape>    short) 404 SB
gr
32 0 0 38 38 0 0 0 34 /Times-Roman /font32 ANSIFont font
gs 826 49 1168 1875 CB
1190 1881 232 ( 'a short pencil') 232 SB
gr
/fm 256 def
2 2 390 1924 B
1 F
n
/fm 256 def
414 2 393 1924 B
1 F
n
/fm 256 def
2 2 808 1924 B
1 F
n
/fm 256 def
353 2 811 1924 B
1 F
n
/fm 256 def
2 2 1165 1924 B
1 F
n
/fm 256 def
825 2 1168 1924 B
1 F
n
/fm 256 def
2 2 1994 1924 B
1 F
n
/fm 256 def
2 186 390 1927 B
1 F
n
/fm 256 def
2 186 808 1927 B
1 F
n
/fm 256 def
2 186 1165 1927 B
1 F
n
/fm 256 def
2 186 1994 1927 B
1 F
n
gs 415 49 393 1924 CB
414 1930 266 (5. Noun modifier) 266 SB
gr
gs 354 49 811 1924 CB
833 1930 88 (CL-N) 88 SB
gr
gs 826 49 1168 1924 CB
1190 1930 313 (/kana   naktongtiew/) 313 SB
gr
gs 826 49 1168 1970 CB
1190 1976 247 (    \(CL\)        \(N\)) 247 SB
gr
gs 826 49 1168 2016 CB
1190 2021 20 (  ) 20 SB
gr
32 0 0 38 38 0 0 0 35 /Times-Italic /font31 ANSIFont font
gs 826 49 1168 2016 CB
1210 2020 232 (group    tourist) 232 SB
gr
32 0 0 38 38 0 0 0 34 /Times-Roman /font32 ANSIFont font
gs 826 49 1168 2062 CB
1190 2068 291 ( 'a group of tourist') 291 SB
gr
/fm 256 def
2 2 390 2111 B
1 F
n
/fm 256 def
414 2 393 2111 B
1 F
n
/fm 256 def
2 2 808 2111 B
1 F
n
/fm 256 def
353 2 811 2111 B
1 F
n
/fm 256 def
2 2 1165 2111 B
1 F
n
/fm 256 def
825 2 1168 2111 B
1 F
n
/fm 256 def
2 2 1994 2111 B
1 F
n
/fm 256 def
2 827 390 2114 B
1 F
n
/fm 256 def
2 2 390 2942 B
1 F
n
/fm 256 def
2 2 390 2942 B
1 F
n
/fm 256 def
414 2 393 2942 B
1 F
n
/fm 256 def
2 827 808 2114 B
1 F
n
/fm 256 def
2 2 808 2942 B
1 F
n
/fm 256 def
353 2 811 2942 B
1 F
n
/fm 256 def
2 827 1165 2114 B
1 F
n
/fm 256 def
2 2 1165 2942 B
1 F
n
/fm 256 def
825 2 1168 2942 B
1 F
n
/fm 256 def
2 827 1994 2114 B
1 F
n
/fm 256 def
2 2 1994 2942 B
1 F
n
/fm 256 def
2 2 1994 2942 B
1 F
n
gs 415 49 393 2111 CB
414 2117 168 (6. Pronoun) 168 SB
gr
gs 415 49 393 2157 CB
414 2163 318 (    -Relative pronoun) 318 SB
gr
gs 415 49 393 2387 CB
414 2393 252 (    -Interrogative) 252 SB
gr
gs 415 49 393 2433 CB
414 2439 177 (     pronoun) 177 SB
gr
gs 415 49 393 2571 CB
414 2577 307 (    -Ordinal pronoun) 307 SB
gr
gs 415 49 393 2755 CB
414 2761 182 (    -Pronoun) 182 SB
gr
gs 354 49 811 2111 CB
833 2117 225 (a\) CL-REL_M) 225 SB
gr
gs 354 49 811 2387 CB
833 2393 217 (b\) CL-ITR_M) 217 SB
gr
gs 354 49 811 2571 CB
833 2577 216 (c\) CL-DONM) 216 SB
gr
gs 354 49 811 2755 CB
833 2761 209 (d\) CL-DDAC) 209 SB
gr
gs 826 49 1168 2111 CB
1190 2117 548 (a\) /nakbanchii  khon thii  thamngan) 548 SB
gr
gs 826 49 1168 2157 CB
1190 2163 614 (             \(N\)            \(CL\) \(REL-M\)   \(V\)) 614 SB
gr
gs 826 49 1168 2203 CB
1190 2208 60 (      ) 60 SB
gr
32 0 0 38 38 0 0 0 35 /Times-Italic /font31 ANSIFont font
gs 826 49 1168 2203 CB
1250 2207 449 (accountant       who       work) 449 SB
gr
32 0 0 38 38 0 0 0 34 /Times-Roman /font32 ANSIFont font
gs 826 49 1168 2249 CB
1190 2255 329 (      thii  borisat    nii/) 329 SB
gr
32 0 0 38 38 0 0 0 35 /Times-Italic /font31 ANSIFont font
gs 826 49 1168 2295 CB
1190 2299 333 (      at   company  this) 333 SB
gr
32 0 0 38 38 0 0 0 34 /Times-Roman /font32 ANSIFont font
gs 826 49 1168 2341 CB
1190 2347 719 (     'the accountant who works at this company') 719 SB
gr
gs 826 49 1168 2387 CB
1190 2393 225 (b\)  /sing    nai/) 225 SB
gr
gs 826 49 1168 2433 CB
1190 2439 328 (       \(CL\)     \(ITR-M\)) 328 SB
gr
gs 826 49 1168 2479 CB
1190 2484 60 (      ) 60 SB
gr
32 0 0 38 38 0 0 0 35 /Times-Italic /font31 ANSIFont font
gs 826 49 1168 2479 CB
1250 2483 232 (<thing> which) 232 SB
gr
32 0 0 38 38 0 0 0 34 /Times-Roman /font32 ANSIFont font
gs 826 49 1168 2525 CB
1190 2531 232 (      'which one') 232 SB
gr
gs 826 49 1168 2571 CB
1190 2577 215 (c\)  / tua  raek/) 215 SB
gr
gs 826 49 1168 2617 CB
1190 2623 305 (       \(CL\)  \(DONM\)) 305 SB
gr
gs 826 49 1168 2663 CB
1190 2668 80 (        ) 80 SB
gr
32 0 0 38 38 0 0 0 35 /Times-Italic /font31 ANSIFont font
gs 826 49 1168 2663 CB
1270 2667 158 (one    first) 158 SB
gr
32 0 0 38 38 0 0 0 34 /Times-Roman /font32 ANSIFont font
gs 826 49 1168 2709 CB
1190 2715 259 (      'the first one') 259 SB
gr
gs 826 49 1168 2755 CB
1190 2761 528 (d\)  /khon   nii   chop   bia      mak/) 528 SB
gr
gs 826 49 1168 2801 CB
1190 2807 306 (       \(CL\)   \(DDAC\)) 306 SB
gr
gs 826 49 1168 2847 CB
1190 2852 40 (    ) 40 SB
gr
32 0 0 38 38 0 0 0 35 /Times-Italic /font31 ANSIFont font
gs 826 49 1168 2847 CB
1230 2851 482 (    the    one  like    beer     very) 482 SB
gr
32 0 0 38 38 0 0 0 34 /Times-Roman /font32 ANSIFont font
gs 826 49 1168 2893 CB
1190 2899 520 (     'The one likes beer very much') 520 SB
gr
315 2948 10 (.) 10 SB
32 0 0 42 42 0 0 0 41 /Times-Bold /font29 ANSIFont font
681 2991 134 (Fig. 2.1) 134 SB
32 0 0 42 42 0 0 0 38 /Times-Roman /font32 ANSIFont font
815 2994 755 ( Classification of classifier expressions table) 755 SB
1 #C
statusdict begin /manualfeed false store end
EJ RS
%%PageTrailer
%%PageResources: font Times-Bold
%%+ font Times-Italic
%%+ font Times-Roman
%%Page: 4 4
%%PageResources: (atend)
SS
0 0 25 31 776 1169 300 SM
0 0 0 fC
/fm 256 def
2 2 369 253 B
1 F
n
/fm 256 def
2 2 369 253 B
1 F
n
/fm 256 def
746 2 372 253 B
1 F
n
/fm 256 def
2 2 1119 253 B
1 F
n
/fm 256 def
746 2 1122 253 B
1 F
n
/fm 256 def
2 2 1869 253 B
1 F
n
/fm 256 def
2 2 1869 253 B
1 F
n
/fm 256 def
2 681 369 256 B
1 F
n
/fm 256 def
2 2 369 938 B
1 F
n
/fm 256 def
2 2 369 938 B
1 F
n
/fm 256 def
746 2 372 938 B
1 F
n
/fm 256 def
2 681 1119 256 B
1 F
n
/fm 256 def
2 2 1119 938 B
1 F
n
/fm 256 def
746 2 1122 938 B
1 F
n
/fm 256 def
2 681 1869 256 B
1 F
n
/fm 256 def
2 2 1869 938 B
1 F
n
/fm 256 def
2 2 1869 938 B
1 F
n

/MSTT31c12c [50.0 0 0 0 0 0] 20 -40 [-50.0 -50.0 50.0 50.0] [1 50 div 0 0 1 50 div 0 0] /MSTT31c12c GreNewFont

32 0 0 50 50 0 0 1 43 /MSTT31c12c font

/G28 [16.0 0.0 4.0 -9.0 14.0 28.0]
/G28 {
    10 37 true [1 0 0 -1 -4.0 28.0] {<018003c0070006000c001c001800300030006000600060006000c000c000c000c000c000c000c000
c000c000c000e00060006000600030003000380018000c000e000700038001c00080>} imagemask 
  }
  40 /G28 MSTT31c12c AddChar
/Ga4 [21.0 0.0 2.0 0.0 18.0 20.0]
/Ga4 {
    16 20 true [1 0 0 -1 -2.0 20.0] {<07e01ff8381e7006e007c003cf83cfc3d8c3d8c3fd837f8378033803380318031803180318031803
>} imagemask 
  }
  164 /Ga4 MSTT31c12c AddChar
/Gb3 [30.0 0.0 2.0 0.0 29.0 20.0]
/Gb3 {
    27 20 true [1 0 0 -1 -2.0 20.0] {<03f806001ffc06003c0e0600f80306001e030600070306000c0306001c0306001803060018030600
180306001803078018031fc018033ee01f8376601983e66018c3c66018c3c6601f8387e00f0303c0
>} imagemask 
  }
  179 /Gb3 MSTT31c12c AddChar
/Gd0 [16.0 0.0 2.0 0.0 14.0 19.0]
/Gd0 {
    12 19 true [1 0 0 -1 -2.0 19.0] {<7810cc10c430c4f0ffc07f801c00000000000000000000007810cc10c430c4f0ffc07f801c00>} imagemask 
  }
  208 /Gd0 MSTT31c12c AddChar
/Ga1 [21.0 0.0 1.0 0.0 18.0 20.0]
/Ga1 {
    17 20 true [1 0 0 -1 -1.0 20.0] {<03f8001ffe003c0700f803801e01800701800c01801c018018018018018018018018018018018018
0180180180180180180180180180180180180180>} imagemask 
  }
  161 /Ga1 MSTT31c12c AddChar
/Gc3 [16.0 0.0 0.0 0.0 16.0 20.0]
/Gc3 {
    16 20 true [1 0 0 -1 0.0 20.0] {<0fc23ffe381f7c00bf8001e00070003800180018001800180018001801f801980118019801f800f0
>} imagemask 
  }
  195 /Gc3 MSTT31c12c AddChar
/Gc1 [22.0 0.0 3.0 0.0 18.0 20.0]
/Gc1 {
    15 20 true [1 0 0 -1 -3.0 20.0] {<7806fc06cc068c06cc06fc060c060c060c060c060c060c067e06fe06cf06cd86ccc6cce6fc7e783c
>} imagemask 
  }
  193 /Gc1 MSTT31c12c AddChar
/Gd2 [17.0 0.0 0.0 0.0 14.0 20.0]
/Gd2 {
    14 20 true [1 0 0 -1 0.0 20.0] {<0fc03ff07838e01cc00c400c000c000c000c000c000c000c000c000c000c000c000c000c000c000c
>} imagemask 
  }
  210 /Gd2 MSTT31c12c AddChar
/G5f [18.0 0.0 2.0 -6.0 17.0 -4.0]
/G5f {
    15 2 true [1 0 0 -1 -2.0 -4.0] {<fffefffe>} imagemask 
  }
  95 /G5f MSTT31c12c AddChar
/G31 [22.0 0.0 5.0 0.0 13.0 23.0]
/G31 {
    8 23 true [1 0 0 -1 -5.0 23.0] {<03071f3b73e30303030303030303030303030303030303>} imagemask 
  }
  49 /G31 MSTT31c12c AddChar
/G2c [10.0 0.0 3.0 -4.0 7.0 4.0]
/G2c {
    4 8 true [1 0 0 -1 -3.0 4.0] {<60f0f0f030302040>} imagemask 
  }
  44 /G2c MSTT31c12c AddChar
/G20 [11.0 0.0 0.0 0.0 0.0 0.0]
/G20 {
} 
  32 /G20 MSTT31c12c AddChar
/G32 [22.0 0.0 3.0 0.0 19.0 23.0]
/G32 {
    16 23 true [1 0 0 -1 -3.0 23.0] {<07e01ff83c1c70067007e003e003e00350070006000e003c007801f007c00f801e003c0038007000
70007fffffff>} imagemask 
  }
  50 /G32 MSTT31c12c AddChar
/G29 [16.0 0.0 3.0 -9.0 13.0 28.0]
/G29 {
    10 37 true [1 0 0 -1 -3.0 28.0] {<c000e000300018001c000e00060007000300038001800180018000c000c000c000c000c000c000c0
00c000c000c001c001800180018003000300060006000c001c0038007000e0004000>} imagemask 
  }
  41 /G29 MSTT31c12c AddChar

gs 747 65 372 253 CB
393 256 505 (\(\244\263\320\241\303\303\301\241\322\303_111, \244\263\320_2, 11\)) 505 SB
gr

/Gc5 [19.0 0.0 2.0 0.0 16.0 20.0]
/Gc5 {
    14 20 true [1 0 0 -1 -2.0 20.0] {<0fc03ff07038e018c00c000c3e0c3f0c638c61cc60ec606c607c603c7e1c661c630c630c7e0c3c0c
>} imagemask 
  }
  197 /Gc5 MSTT31c12c AddChar
/Gd8 [0.0 0.0 -9.0 -13.0 -3.0 -3.0]
/Gd8 {
    6 10 true [1 0 0 -1 9.0 -3.0] {<78fc8c8cfc7c0c0c0c0c>} imagemask 
  }
  216 /Gd8 MSTT31c12c AddChar
/G8b [0.0 0.0 -6.0 23.0 -4.0 29.0]
/G8b {
    2 6 true [1 0 0 -1 6.0 29.0] {<c0c0c0c0c0c0>} imagemask 
  }
  139 /G8b MSTT31c12c AddChar
/G35 [22.0 0.0 3.0 0.0 19.0 23.0]
/G35 {
    16 23 true [1 0 0 -1 -3.0 23.0] {<3ffe3ffe3800380038003000700073f07ff87ffc7c0e7806000700030003000300032003f0067006
3c1c1ff807e0>} imagemask 
  }
  53 /G35 MSTT31c12c AddChar

gs 747 65 372 315 CB
393 318 368 (\(\244\263\320\241\303\303\301\241\322\303_111, \241\305\330\213 ) 357 SB
750 318 121 (\301_2, 5\)) 121 SB
gr

/Gb9 [22.0 0.0 2.0 0.0 21.0 20.0]
/Gb9 {
    19 20 true [1 0 0 -1 -2.0 20.0] {<780600fc0600cc06008c0600cc0600fc06000c06000c06000c06000c06000c06000c0f800c3fc00c
fe600de6600f86600f06600e06600e07e00c03c0>} imagemask 
  }
  185 /Gb9 MSTT31c12c AddChar
/G36 [22.0 0.0 3.0 0.0 19.0 23.0]
/G36 {
    16 23 true [1 0 0 -1 -3.0 23.0] {<03f00ffc1c1e38043000700073f07ff8fe1cfc0ef806f007f003f00370037003700370033806380e
1e1c0ff803e0>} imagemask 
  }
  54 /G36 MSTT31c12c AddChar

gs 747 65 372 377 CB
393 380 459 (\(\244\263\320\241\303\303\301\241\322\303_111, \244\271_1, 6\)) 459 SB
gr

/G33 [22.0 0.0 3.0 0.0 19.0 23.0]
/G33 {
    16 23 true [1 0 0 -1 -3.0 23.0] {<07e01ff83c1c78067006700300030006000e00fc03f803fc001e0007000700030003f00370037806
3c1c1ff807e0>} imagemask 
  }
  51 /G33 MSTT31c12c AddChar
/Gb5 [21.0 0.0 2.0 0.0 18.0 20.0]
/Gb5 {
    16 20 true [1 0 0 -1 -2.0 20.0] {<06181ffc39e670c76003c003c7c3c663cc23cc23666363e331c331831b831b030f030e030e030e03
>} imagemask 
  }
  181 /Gb5 MSTT31c12c AddChar
/Gd1 [0.0 0.0 -11.0 25.0 4.0 32.0]
/Gd1 {
    15 7 true [1 0 0 -1 11.0 32.0] {<7802cc02c406c41efc7c7ff03fc0>} imagemask 
  }
  209 /Gd1 MSTT31c12c AddChar
/Gc7 [18.0 0.0 1.0 0.0 15.0 20.0]
/Gc7 {
    14 20 true [1 0 0 -1 -1.0 20.0] {<0fc03ff07038e01cc00c400c000c000c000c000c000c000c000c000c00fc00cc008c00cc00fc0078
>} imagemask 
  }
  199 /Gc7 MSTT31c12c AddChar
/G39 [22.0 0.0 4.0 0.0 19.0 23.0]
/G39 {
    15 23 true [1 0 0 -1 -4.0 23.0] {<0fc01ff038786018601cc00cc00ec00ec00ec00ec00e601e603e387e1ffe0fce000c000c401c6038
70783ff00fc0>} imagemask 
  }
  57 /G39 MSTT31c12c AddChar

gs 747 65 372 439 CB
393 442 240 (\(\271\241_13111, \265\321 ) 229 SB
622 442 117 (\307_1, 9\)) 117 SB
gr

/Gbd [21.0 0.0 3.0 0.0 18.0 30.0]
/Gbd {
    15 30 true [1 0 0 -1 -3.0 30.0] {<000600060006000600060006000600060006000678067c06c406c406cc06fc06c006c006c006c006
c386c7c6cee6dc66d836f03ef01ee00ee00ec006>} imagemask 
  }
  189 /Gbd MSTT31c12c AddChar
/Gd9 [0.0 0.0 -15.0 -14.0 -4.0 -3.0]
/Gd9 {
    11 11 true [1 0 0 -1 15.0 -3.0] {<3060f860cc608c60cc60fc600c600c600c600fe00fe0>} imagemask 
  }
  217 /Gd9 MSTT31c12c AddChar
/Ga7 [16.0 0.0 1.0 0.0 13.0 20.0]
/Ga7 {
    12 20 true [1 0 0 -1 -1.0 20.0] {<01e003f002300230033001f00030003000300030e030f03038301c300c300e300630073003e001c0
>} imagemask 
  }
  167 /Ga7 MSTT31c12c AddChar
/G34 [22.0 0.0 3.0 0.0 19.0 23.0]
/G34 {
    16 23 true [1 0 0 -1 -3.0 23.0] {<0038007800f800f801d803d8039807180f180e181c183c1838187018f018e018ffffffff00180018
001800180018>} imagemask 
  }
  52 /G34 MSTT31c12c AddChar

gs 747 65 372 501 CB
393 504 240 (\(\271\241_13111, \275\331 ) 229 SB
622 504 115 (\247_2, 4\)) 115 SB
gr

/Ge4 [15.0 0.0 -1.0 0.0 14.0 35.0]
/Ge4 {
    15 35 true [1 0 0 -1 1.0 35.0] {<c1c0e3e0736033603360336036603660366036601c601c6000600060006000600060006000600060
006000600060006000600060006000600060007e006600620062007e003c>} imagemask 
  }
  228 /Ge4 MSTT31c12c AddChar
/G30 [22.0 0.0 3.0 0.0 20.0 23.0]
/G30 {
    17 23 true [1 0 0 -1 -3.0 23.0] {<03f0000ff8001c1c00380600300700600300600300600180c00180c00180c00180c00180c00180c0
0180c00180e001806003006003007007003806001c1c000ff80003f000>} imagemask 
  }
  48 /G30 MSTT31c12c AddChar

gs 747 65 372 563 CB
393 566 63 (\(\344\241\213 ) 52 SB
445 566 181 (_13111, \265\321 ) 170 SB
615 566 139 (\307_1, 10\)) 139 SB
gr

/Ge0 [11.0 0.0 3.0 0.0 9.0 20.0]
/Ge0 {
    6 20 true [1 0 0 -1 -3.0 20.0] {<c0c0c0c0c0c0c0c0c0c0c0c0c0c0fcccc4c4fc78>} imagemask 
  }
  224 /Ge0 MSTT31c12c AddChar
/G8c [0.0 0.0 -10.0 23.0 2.0 31.0]
/G8c {
    12 8 true [1 0 0 -1 10.0 31.0] {<7810c810cc10cc307ce019c01f807c00>} imagemask 
  }
  140 /G8c MSTT31c12c AddChar

gs 747 65 372 625 CB
393 628 63 (\(\344\241\213 ) 52 SB
445 628 190 (_13111, \340\305\214 ) 179 SB
624 628 116 (\322_2, 3\)) 116 SB
gr

/Ga8 [19.0 0.0 1.0 0.0 15.0 20.0]
/Ga8 {
    14 20 true [1 0 0 -1 -1.0 20.0] {<0fc03ff07838e01cc00c000c060c1f0c198c118c198c1f8c0f8c018c018c018c018c018c01f800f0
>} imagemask 
  }
  168 /Ga8 MSTT31c12c AddChar
/Gcd [20.0 0.0 3.0 0.0 17.0 20.0]
/Gcd {
    14 20 true [1 0 0 -1 -3.0 20.0] {<0fc03ff07838601cc00c000c180c3e0c660c630c660c7e0c7c0c600c600c600c600c600c7ffc3ff8
>} imagemask 
  }
  205 /Gcd MSTT31c12c AddChar
/G37 [22.0 0.0 4.0 0.0 19.0 23.0]
/G37 {
    15 23 true [1 0 0 -1 -4.0 23.0] {<fffefffe000e001c001c00380070007000e000e001c001c001800380038003800700070007000700
070007000700>} imagemask 
  }
  55 /G37 MSTT31c12c AddChar

gs 747 65 372 687 CB
393 690 353 (\(\271\241\241\303\320\250\315\241_13111, \265\321 ) 342 SB
735 690 117 (\307_1, 7\)) 117 SB
gr
gs 747 65 372 749 CB
393 752 328 (\(\244\271_111, \244\271_1, 67\)) 328 SB
gr
gs 747 65 372 811 CB
393 814 215 (\(\244\271_111, \241\305\330\213 ) 204 SB
597 814 121 (\301_2, 1\)) 121 SB
gr

/Gb7 [23.0 0.0 2.0 0.0 20.0 20.0]
/Gb7 {
    18 20 true [1 0 0 -1 -2.0 20.0] {<780f00fc1f80cc39c08c70c0cc60c0fce0c00cc0c00d80c00d80c00f00c00f00c00f00c00e00c00e
00c00e00c00c00c00c00c00c00c00c00c00c00c0>} imagemask 
  }
  183 /Gb7 MSTT31c12c AddChar
/Gcb [23.0 0.0 2.0 0.0 20.0 20.0]
/Gcb {
    18 20 true [1 0 0 -1 -2.0 20.0] {<781e00fc3f00cc31808c3180cc3980fc1f000c0f000c1f800c39800c70c00ce0c00cc0c00dc0c00f
80c00f00c00f00c00e00c00e00c00c00c00c00c0>} imagemask 
  }
  203 /Gcb MSTT31c12c AddChar

gs 747 65 372 873 CB
393 876 364 (\(\267\313\322\303_111, \244\271_1, 17\)) 364 SB
gr

/Gc2 [20.0 0.0 3.0 0.0 17.0 20.0]
/Gc2 {
    14 20 true [1 0 0 -1 -3.0 20.0] {<780c7c0ccc0cc40ccc0cfc0ce00cc00cc00ce00c780c3c0c600ce00cc00cc00cc00cc01cfff8fff0
>} imagemask 
  }
  194 /Gc2 MSTT31c12c AddChar

gs 747 65 1122 253 CB
1143 256 358 (\(\267\313\322\303_111, \271\322\302_1, 9\)) 358 SB
gr

/G86 [0.0 0.0 -11.0 23.0 -9.0 29.0]
/G86 {
    2 6 true [1 0 0 -1 11.0 29.0] {<c0c0c0c0c0c0>} imagemask 
  }
  134 /G86 MSTT31c12c AddChar

gs 747 65 1122 315 CB
1143 318 232 (\(\267\313\322\303_111, \275\206 ) 221 SB
1364 318 136 (\322\302_2, 1\)) 136 SB
gr
gs 747 65 1122 377 CB
1143 380 361 (\(\244\271\247\322\271_111, \244\271_1, 6\)) 361 SB
gr

/Gca [20.0 0.0 3.0 0.0 20.0 20.0]
/Gca {
    17 20 true [1 0 0 -1 -3.0 20.0] {<0fc1003ff380703f00e01e00c01c00003c003eec003fcc00638c0061cc0060ec00606c00607c0060
3c007e1c00661c00620c00630c007e0c003c0c00>} imagemask 
  }
  202 /Gca MSTT31c12c AddChar

gs 747 65 1122 439 CB
1143 442 47 (\(\312\214 ) 36 SB
1179 442 201 (\301_13114, \305\331 ) 190 SB
1369 442 142 (\241_1, 12\)) 142 SB
gr

/Gbc [21.0 0.0 3.0 0.0 18.0 20.0]
/Gbc {
    15 20 true [1 0 0 -1 -3.0 20.0] {<78067c06c406c406cc06fc06c006c006c006c006c386c7c6cee6dc66d836f03ef01ee00ee00ec006
>} imagemask 
  }
  188 /Gbc MSTT31c12c AddChar

gs 747 65 1122 501 CB
1143 504 47 (\(\312\214 ) 36 SB
1179 504 310 (\301_13114, \274\305_1, 3\)) 310 SB
gr

/Ge1 [19.0 0.0 3.0 0.0 18.0 20.0]
/Ge1 {
    15 20 true [1 0 0 -1 -3.0 20.0] {<c060c060c060c060c060c060c060c060c060c060c060c060c060c060fc7ecc66c662c6627c7e783c
>} imagemask 
  }
  225 /Ge1 MSTT31c12c AddChar
/Ge2 [14.0 0.0 -1.0 0.0 13.0 35.0]
/Ge2 {
    14 35 true [1 0 0 -1 1.0 35.0] {<1f003ff8707cfc000e00030001800180018001800180018001800180018001800180018001800180
01800180018001800180018001800180018001f801980188018801f800f0>} imagemask 
  }
  226 /Ge2 MSTT31c12c AddChar
/G38 [22.0 0.0 3.0 0.0 18.0 23.0]
/G38 {
    15 23 true [1 0 0 -1 -3.0 23.0] {<0fe03ff8783c701c600ce00c600c701c78381ff00ff03838701ce00ee00ee00ee00ee00ee00e701c
783c3ff80fe0>} imagemask 
  }
  56 /G38 MSTT31c12c AddChar

gs 747 65 1122 563 CB
1143 566 287 (\(\341\265\247\342\301_13114, \305\331 ) 276 SB
1419 566 120 (\241_1, 8\)) 120 SB
gr

/Gd5 [0.0 0.0 -18.0 23.0 -4.0 33.0]
/Gd5 {
    14 10 true [1 0 0 -1 18.0 33.0] {<001c001c0f9c1ffc387c601cffccfffce01c0004>} imagemask 
  }
  213 /Gd5 MSTT31c12c AddChar

gs 747 65 1122 625 CB
1143 628 50 (\(\267\330 ) 39 SB
1182 628 38 (\340\303\325 ) 27 SB
1209 628 221 (\302\271_13114, \305\331 ) 210 SB
1419 628 120 (\241_1, 9\)) 120 SB
gr
gs 747 65 1122 687 CB
1143 690 232 (\(\342\244_13111, \265\321 ) 221 SB
1364 690 117 (\307_1, 7\)) 117 SB
gr
gs 747 65 1122 749 CB
1143 752 259 (\(\313\301\322_13111, \265\321 ) 248 SB
1391 752 139 (\307_1, 13\)) 139 SB
gr
gs 747 65 1122 811 CB
1143 814 72 (\(\313\301\331 ) 61 SB
1204 814 181 (_13111, \265\321 ) 170 SB
1374 814 117 (\307_1, 5\)) 117 SB
gr

/Gaa [21.0 0.0 1.0 0.0 20.0 22.0]
/Gaa {
    19 22 true [1 0 0 -1 -1.0 22.0] {<0000400000601f00e07fc3c0fcff00c67c00c67f00c67300fe61803c618000c18000c18000c18000
c18000c18000c18000c18000c18000c18000c18003ff0003fe00>} imagemask 
  }
  170 /Gaa MSTT31c12c AddChar
/Gd7 [0.0 0.0 -18.0 23.0 -4.0 33.0]
/Gd7 {
    14 10 true [1 0 0 -1 18.0 33.0] {<01d801d80fd83ff878786018ffc8fffce01c0004>} imagemask 
  }
  215 /Gd7 MSTT31c12c AddChar

gs 747 65 1122 873 CB
1143 876 48 (\(\252\214 ) 37 SB
1180 876 225 (\322\247_13111, \340\252\327 ) 214 SB
1394 876 140 (\315\241_1, 3\)) 140 SB
gr
32 0 0 42 42 0 0 0 41 /Times-Bold /font29 ANSIFont font
706 1038 134 (Fig. 3.1) 134 SB
32 0 0 42 42 0 0 0 38 /Times-Roman /font32 ANSIFont font
840 1041 784 ( Table of Noun Classifier Associations \(NCA\)) 784 SB
32 0 0 34 34 0 0 0 34 /AvantGarde-Book /font1 ANSIFont font
gs 1798 732 165 1281 CB
647 1281 213 (Concrete \(1\)) 213 SB
gr
1 lc
1 lj
0 0 0 pC
6 3 SP
458 1453 M 258 -120 1 PP
S
n
gs 1798 732 165 1281 CB
372 1454 197 (Subject \(11\)) 197 SB
gr
458 1505 M -190 120 1 PP
S
n
gs 1798 732 165 1281 CB
165 1626 200 (Person \(111\)) 200 SB
gr
716 1333 M 241 120 1 PP
S
n
gs 1798 732 165 1281 CB
768 1454 338 (Concrete place \(12\)) 338 SB
gr
458 1505 M 189 120 1 PP
S
n
gs 1798 732 165 1281 CB
423 1626 305 (Organization \(112\)) 305 SB
gr
975 1505 M -190 120 1 PP
S
n
gs 1798 732 165 1281 CB
768 1626 27 (...) 27 SB
gr
975 1505 M 189 120 1 PP
S
n
gs 1798 732 165 1281 CB
1147 1626 27 (...) 27 SB
gr
716 1333 M 724 120 1 PP
S
n
gs 1798 732 165 1281 CB
1354 1454 325 (Concrete thing \(13\)) 325 SB
gr
1491 1505 M -189 120 1 PP
S
n
gs 1798 732 165 1281 CB
1233 1626 298 (Nature thing \(131\)) 298 SB
gr
gs 1798 732 165 1281 CB
1681 1626 27 (...) 27 SB
gr
1491 1505 M 414 120 1 PP
S
n
1491 1505 M 190 120 1 PP
S
n
gs 1798 732 165 1281 CB
1888 1626 27 (...) 27 SB
gr
1302 1677 M -190 121 1 PP
S
n
gs 1798 732 165 1281 CB
1009 1798 297 (Living thing \(1311\)) 297 SB
gr
1095 1849 M -189 121 1 PP
S
n
1302 1677 M 189 121 1 PP
S
n
1095 1849 M 190 121 1 PP
S
n
1302 1677 M 413 121 1 PP
S
n
1095 1849 M 414 121 1 PP
S
n
gs 1798 732 165 1281 CB
837 1970 245 (Animal \(13111\)) 245 SB
gr
gs 1798 732 165 1281 CB
1474 1798 27 (...) 27 SB
gr
gs 1798 732 165 1281 CB
1698 1798 27 (...) 27 SB
gr
gs 1798 732 165 1281 CB
1267 1970 27 (...) 27 SB
gr
gs 1798 732 165 1281 CB
1457 1970 213 (Plant \(13113\)) 213 SB
gr
1095 1849 M 724 121 1 PP
S
n
gs 1798 732 165 1281 CB
1750 1970 196 (Fruit \(13114\)) 196 SB
gr
32 0 0 42 42 0 0 0 41 /Times-Bold /font29 ANSIFont font
937 2165 134 (Fig. 3.2) 134 SB
32 0 0 42 42 0 0 0 38 /Times-Roman /font32 ANSIFont font
1071 2168 323 ( Concept hierarchy) 323 SB
165 2269 133 (Instead ) 140 SB
305 2269 46 (of ) 53 SB
358 2269 138 (picking ) 145 SB
503 2269 53 (up ) 60 SB
563 2269 63 (the ) 70 SB
633 2269 82 (data ) 90 SB
723 2269 157 (sentence ) 165 SB
888 2269 53 (by ) 61 SB
949 2269 157 (sentence,) 157 SB
165 2319 60 (we ) 61 SB
226 2319 167 (extracted ) 168 SB
394 2319 30 (a ) 31 SB
425 2319 164 (fragment ) 165 SB
590 2319 46 (of ) 48 SB
638 2319 82 (data ) 84 SB
722 2319 128 (around ) 130 SB
852 2319 63 (the ) 65 SB
32 0 0 42 42 0 0 0 39 /Times-Italic /font31 ANSIFont font
917 2318 31 (cl) 31 SB
32 0 0 42 42 0 0 0 38 /Times-Roman /font32 ANSIFont font
948 2319 22 (, ) 24 SB
972 2319 134 (because) 134 SB
165 2371 96 (there ) 117 SB
282 2371 39 (is ) 60 SB
342 2371 53 (no ) 74 SB
416 2371 139 (explicit ) 160 SB
576 2371 131 (marker ) 152 SB
728 2371 44 (to ) 65 SB
793 2371 146 (indicate ) 167 SB
960 2371 146 (sentence) 146 SB
165 2421 207 (boundaries. ) 217 SB
382 2421 70 (We ) 80 SB
462 2421 88 (used ) 98 SB
560 2421 63 (the ) 73 SB
633 2421 105 (range ) 115 SB
748 2421 46 (of ) 56 SB
804 2421 67 (-10 ) 78 SB
882 2421 44 (to ) 55 SB
937 2421 56 (+2 ) 67 SB
1004 2421 102 (words) 102 SB
165 2471 128 (around ) 136 SB
301 2471 63 (the ) 71 SB
32 0 0 42 42 0 0 0 39 /Times-Italic /font31 ANSIFont font
372 2470 42 (cl ) 50 SB
32 0 0 42 42 0 0 0 38 /Times-Roman /font32 ANSIFont font
422 2471 44 (in ) 52 SB
474 2471 67 (our ) 76 SB
550 2471 218 (experiments ) 227 SB
777 2471 114 (which ) 123 SB
900 2471 164 (appeared ) 173 SB
1073 2471 33 (to) 33 SB
165 2523 640 (cover most of co-occurrence patterns.) 640 SB
32 0 0 42 42 0 0 0 41 /Times-Bold /font29 ANSIFont font
165 2620 111 (Step 4) 111 SB
32 0 0 42 42 0 0 0 38 /Times-Roman /font32 ANSIFont font
276 2623 12 (:) 12 SB
315 2623 289 (Pattern matching) 289 SB
165 2674 101 (Input:) 101 SB
315 2674 285 (Output of step 3.) 285 SB
165 2724 129 (Output:) 129 SB
315 2724 699 (A list of nouns-classifiers with frequency) 699 SB
165 2774 524 (information of co-occurrences.) 524 SB
165 2874 876 (In this step, the tagged corpus is matched with each) 876 SB
165 2924 787 (pattern of classifier occurrences shown below:) 787 SB
315 3024 520 (N- -NCNM-CL \(Enumeration\)) 520 SB
1360 2275 148 (N- -CL- ) 148 SB
32 0 0 50 50 0 0 1 47 /MSTT31c146 font

/Gb7 [22.0 0.0 1.0 0.0 20.0 20.0]
/Gb7 {
    19 20 true [1 0 0 -1 -1.0 20.0] {<3c07806e0fc0471be0c711e06721e03f21e00721e00721e00721e00741e00741e00741e00741e007
81e00781e00781e00781e00781e00701e00701e0>} imagemask 
  }
  183 /Gb7 MSTT31c146 AddChar

1508 2266 33 (\267\325\350 ) 22 SB
1530 2266 11 ( ) 11 SB
32 0 0 42 42 0 0 0 38 /Times-Roman /font32 ANSIFont font
1541 2275 558 (/tii/-NCNM \(Ordinal expression\)) 558 SB
1360 2336 638 (N- -CL-DET \(Referential expression\)) 638 SB
1360 2386 219 (N- -DET-CL) 219 SB
1510 2436 637 (\(Indefinite demonstration expression\)) 637 SB
1360 2486 657 (N- -CL-VATT \(Attribute noun phrase\)) 657 SB
1360 2536 387 (CL-N \(Noun modifier\)) 387 SB
1360 2592 157 (N- -CL-{) 157 SB
32 0 0 50 50 0 0 1 47 /MSTT31c146 font
1517 2583 33 (\267\325\350 ) 22 SB
1539 2583 11 ( ) 11 SB
32 0 0 42 42 0 0 0 38 /Times-Roman /font32 ANSIFont font
1550 2592 82 (/tii/, ) 82 SB
32 0 0 50 50 0 0 1 47 /MSTT31c146 font

/Gab [22.0 0.0 1.0 0.0 21.0 20.0]
/Gab {
    20 20 true [1 0 0 -1 -1.0 20.0] {<10c07039e0f03bf0e06670e04471c0407f80c07e004c73004473806671c03c71c01871c00071c000
71c00071c00071c000718000738000738003ff00>} imagemask 
  }
  171 /Gab MSTT31c146 AddChar

1632 2583 33 (\253\326\350 ) 22 SB
1654 2583 27 (\247 ) 27 SB
32 0 0 42 42 0 0 0 38 /Times-Roman /font32 ANSIFont font
1681 2592 125 (/sung/, ) 125 SB
32 0 0 50 50 0 0 1 47 /MSTT31c146 font

/Ge3 [15.0 0.0 1.0 0.0 14.0 33.0]
/Ge3 {
    13 33 true [1 0 0 -1 -1.0 33.0] {<0e003f8063c041c049c0d5c045c045c039c001c001c001c001c001c001c001c001c001c001c001c0
01c001c001c001c001c001c001c001e0011002100318011000e0>} imagemask 
  }
  227 /Ge3 MSTT31c146 AddChar

1806 2583 49 (\343\271 ) 49 SB
32 0 0 42 42 0 0 0 38 /Times-Roman /font32 ANSIFont font
1855 2592 129 (/nai/,..}) 129 SB
1510 2653 563 (\(Relative/ Interrogative pronoun\)) 563 SB
1210 2753 114 (where ) 119 SB
1329 2753 41 (N ) 46 SB
1375 2753 140 (denotes ) 145 SB
1520 2753 106 (noun, ) 111 SB
1631 2753 65 (CL ) 70 SB
1701 2753 140 (denotes ) 145 SB
1846 2753 175 (classifier, ) 180 SB
2026 2753 125 (NCNM) 125 SB
1210 2803 140 (denotes ) 154 SB
1364 2803 148 (cardinal ) 163 SB
1527 2803 151 (number, ) 166 SB
1693 2803 93 (DET ) 108 SB
1801 2803 140 (denotes ) 155 SB
1956 2803 195 (determiner,) 195 SB
1210 2859 449 (VATT denotes attributive ) 450 SB
1660 2859 97 (verb, ) 98 SB
32 0 0 50 50 0 0 1 47 /MSTT31c146 font
1758 2850 33 (\267\325\350 ) 22 SB
1780 2850 11 ( ) 12 SB
32 0 0 42 42 0 0 0 38 /Times-Roman /font32 ANSIFont font
1792 2859 82 (/tii/, ) 83 SB
32 0 0 50 50 0 0 1 47 /MSTT31c146 font
1875 2850 33 (\253\326\350 ) 22 SB
1897 2850 27 (\247 ) 28 SB
32 0 0 42 42 0 0 0 38 /Times-Roman /font32 ANSIFont font
1925 2859 114 (/sung/ ) 115 SB
2040 2859 72 (and ) 73 SB
32 0 0 50 50 0 0 1 47 /MSTT31c146 font
2113 2850 38 (\343\271) 38 SB
32 0 0 42 42 0 0 0 38 /Times-Roman /font32 ANSIFont font
1210 2920 87 (/nai/ ) 114 SB
1324 2920 63 (are ) 90 SB
1414 2920 143 (specific ) 170 SB
1584 2920 89 (Thai ) 117 SB
1701 2920 124 (words, ) 152 SB
1853 2920 83 (A-B ) 111 SB
1964 2920 140 (denotes ) 168 SB
2132 2920 19 (a) 19 SB
1210 2970 211 (consecutive ) 223 SB
1433 2970 77 (pair ) 90 SB
1523 2970 46 (of ) 59 SB
1582 2970 41 (A ) 54 SB
1636 2970 72 (and ) 85 SB
1721 2970 50 (B, ) 63 SB
1784 2970 72 (and ) 85 SB
1869 2970 97 (A--B ) 110 SB
1979 2970 140 (denotes ) 153 SB
2132 2970 19 (a) 19 SB
1210 3020 151 (possibly ) 177 SB
1387 3020 171 (separated ) 197 SB
1584 3020 88 (pair. ) 114 SB
1698 3020 168 (Actually, ) 194 SB
1892 3020 97 (A--B ) 123 SB
2015 3020 70 (can ) 96 SB
2111 3020 40 (be) 40 SB
1 #C
statusdict begin /manualfeed false store end
EJ RS
%%PageTrailer
%%PageResources: font AvantGarde-Book
%%+ font MSTT31c12c
%%+ font MSTT31c146
%%+ font Times-Bold
%%+ font Times-Italic
%%+ font Times-Roman
%%Page: 5 5
%%PageResources: (atend)
SS
0 0 25 31 776 1169 300 SM
32 0 0 42 42 0 0 0 38 /Times-Roman /font32 ANSIFont font
0 0 0 fC
165 210 171 (separated ) 192 SB
357 210 53 (by ) 74 SB
431 210 131 (several ) 152 SB
583 210 157 (arbitrary ) 179 SB
762 210 113 (words ) 135 SB
897 210 65 (but ) 87 SB
984 210 44 (in ) 66 SB
1050 210 56 (our) 56 SB
165 260 218 (experiments ) 226 SB
391 260 60 (we ) 69 SB
460 260 194 (considered ) 203 SB
663 260 86 (only ) 95 SB
758 260 149 (possible ) 158 SB
916 260 190 (separations) 190 SB
165 310 53 (by ) 60 SB
225 310 30 (a ) 37 SB
262 310 139 (relative ) 146 SB
408 310 151 (pronoun ) 158 SB
566 310 121 (phrase ) 128 SB
694 310 126 (having ) 133 SB
827 310 53 (no ) 60 SB
887 310 98 (more ) 106 SB
993 310 84 (than ) 92 SB
1085 310 21 (5) 21 SB
165 360 124 (words. ) 134 SB
299 360 86 (This ) 97 SB
396 360 39 (is ) 50 SB
446 360 44 (to ) 55 SB
501 360 92 (limit ) 103 SB
604 360 63 (the ) 74 SB
678 360 119 (search ) 130 SB
808 360 105 (space ) 116 SB
924 360 46 (of ) 57 SB
981 360 125 (general) 125 SB
165 410 100 (cases ) 122 SB
287 410 44 (to ) 66 SB
353 410 30 (a ) 52 SB
405 410 214 (manageable ) 236 SB
641 410 77 (size ) 99 SB
740 410 86 (with ) 109 SB
849 410 100 (some ) 123 SB
972 410 76 (loss ) 99 SB
1071 410 35 (of) 35 SB
165 460 181 (generality.) 181 SB
315 510 77 (The ) 80 SB
395 510 129 (pattern ) 132 SB
527 510 169 (matching ) 172 SB
699 510 137 (process ) 140 SB
839 510 76 (was ) 80 SB
919 510 129 (carried ) 133 SB
1052 510 54 (out) 54 SB
165 560 72 (one ) 74 SB
239 560 53 (by ) 55 SB
294 560 72 (one ) 74 SB
368 560 86 (with ) 88 SB
456 560 89 (each ) 91 SB
547 560 140 (pattern. ) 142 SB
689 560 69 (For ) 72 SB
761 560 89 (each ) 92 SB
853 560 129 (pattern ) 132 SB
985 560 46 (of ) 49 SB
1034 560 55 (A- ) 58 SB
1092 560 14 (-) 14 SB
165 610 92 (B-C, ) 100 SB
265 610 63 (the ) 71 SB
336 610 169 (matching ) 177 SB
513 610 46 (of ) 54 SB
567 610 81 (B-C ) 89 SB
656 610 77 (pair ) 86 SB
742 610 76 (was ) 85 SB
827 610 124 (simple ) 133 SB
960 610 72 (and ) 81 SB
1041 610 65 (was) 65 SB
165 660 187 (performed ) 190 SB
355 660 42 (at ) 45 SB
400 660 90 (first. ) 93 SB
493 660 104 (Next, ) 107 SB
600 660 63 (the ) 67 SB
667 660 169 (matching ) 173 SB
840 660 46 (of ) 50 SB
890 660 30 (a ) 34 SB
924 660 77 (pair ) 81 SB
1005 660 55 (A- ) 59 SB
1064 660 42 (-B) 42 SB
165 710 223 (was done by:) 223 SB
315 810 43 (1. ) 56 SB
371 810 173 (searching ) 186 SB
557 810 60 (for ) 73 SB
630 810 63 (the ) 77 SB
707 810 131 (nearest ) 145 SB
852 810 41 (A ) 55 SB
907 810 93 (from ) 107 SB
1014 810 50 (B. ) 64 SB
1078 810 28 (If) 28 SB
165 860 280 (found, mark A1.) 280 SB
315 910 43 (2. ) 45 SB
360 910 93 (from ) 95 SB
455 910 39 (B ) 41 SB
496 910 119 (within ) 121 SB
617 910 30 (a ) 33 SB
650 910 88 (span ) 91 SB
741 910 46 (of ) 49 SB
790 910 88 (five, ) 91 SB
881 910 173 (searching ) 176 SB
1057 910 49 (for) 49 SB
165 960 63 (the ) 64 SB
229 960 131 (nearest ) 132 SB
361 960 139 (relative ) 140 SB
501 960 162 (pronoun. ) 163 SB
664 960 39 (If ) 40 SB
704 960 120 (found, ) 121 SB
825 960 98 (mark ) 99 SB
924 960 53 (p1 ) 54 SB
978 960 84 (then ) 86 SB
1064 960 42 (go) 42 SB
165 1010 459 (to 3. Otherwise, match A1.) 459 SB
315 1060 43 (3. ) 44 SB
359 1060 126 (further ) 127 SB
486 1060 173 (searching ) 174 SB
660 1060 60 (for ) 61 SB
721 1060 63 (the ) 64 SB
785 1060 131 (nearest ) 132 SB
917 1060 41 (A ) 42 SB
959 1060 93 (from ) 94 SB
1053 1060 53 (p1.) 53 SB
165 1110 39 (If ) 45 SB
210 1110 120 (found, ) 126 SB
336 1110 98 (mark ) 104 SB
440 1110 73 (A2. ) 79 SB
519 1110 39 (If ) 45 SB
564 1110 62 (A2 ) 68 SB
632 1110 39 (is ) 45 SB
677 1110 124 (farther ) 130 SB
807 1110 93 (from ) 100 SB
907 1110 39 (B ) 46 SB
953 1110 84 (than ) 91 SB
1044 1110 62 (A1,) 62 SB
165 1160 560 (match A2. Otherwise, match A1.) 560 SB
315 1260 188 (At the end ) 189 SB
504 1260 46 (of ) 47 SB
551 1260 98 (these ) 99 SB
650 1260 106 (steps, ) 107 SB
757 1260 60 (we ) 61 SB
818 1260 157 (obtained ) 158 SB
976 1260 30 (a ) 31 SB
1007 1260 63 (list ) 64 SB
1071 1260 35 (of) 35 SB
165 1310 111 (nouns ) 113 SB
32 0 0 42 42 0 0 0 39 /Times-Italic /font31 ANSIFont font
278 1309 51 (Ni ) 53 SB
32 0 0 42 42 0 0 0 38 /Times-Roman /font32 ANSIFont font
331 1310 11 ( ) 13 SB
344 1310 105 (along ) 108 SB
452 1310 86 (with ) 89 SB
541 1310 63 (the ) 66 SB
607 1310 180 (frequency ) 183 SB
790 1310 46 (of ) 49 SB
32 0 0 42 42 0 0 0 39 /Times-Italic /font31 ANSIFont font
839 1309 28 (w) 28 SB
32 0 0 42 42 0 0 0 38 /Times-Roman /font32 ANSIFont font
867 1310 11 ( ) 14 SB
881 1310 44 (in ) 47 SB
928 1310 63 (the ) 66 SB
994 1310 112 (corpus) 112 SB
165 1362 60 (for ) 71 SB
236 1362 89 (each ) 100 SB
336 1362 169 (matching ) 180 SB
516 1362 129 (pattern ) 140 SB
656 1362 79 (\(see ) 90 SB
32 0 0 42 42 0 0 0 41 /Times-Bold /font29 ANSIFont font
746 1359 81 (Fig. ) 92 SB
838 1359 53 (3.1) 53 SB
32 0 0 42 42 0 0 0 38 /Times-Roman /font32 ANSIFont font
891 1362 11 ( ) 23 SB
914 1362 60 (for ) 72 SB
986 1362 120 (sample) 120 SB
165 1413 160 (outputs\). ) 167 SB
332 1413 96 (Each ) 103 SB
435 1413 98 (entry ) 105 SB
540 1413 39 (is ) 47 SB
587 1413 46 (of ) 54 SB
641 1413 63 (the ) 71 SB
712 1413 93 (form ) 101 SB
813 1413 148 (\(W_N1, ) 156 SB
969 1413 137 (CL_N2,) 137 SB
165 1463 102 (Freq\) ) 106 SB
271 1463 114 (where ) 118 SB
389 1463 51 (W ) 55 SB
444 1463 140 (denotes ) 144 SB
588 1463 30 (a ) 34 SB
622 1463 106 (noun, ) 110 SB
732 1463 62 (N1 ) 66 SB
798 1463 140 (denotes ) 144 SB
942 1463 30 (a ) 35 SB
977 1463 129 (number) 129 SB
165 1513 220 (representing ) 234 SB
399 1513 162 (semantic ) 176 SB
575 1513 93 (class ) 107 SB
682 1513 46 (of ) 60 SB
742 1513 62 (W, ) 77 SB
819 1513 65 (CL ) 80 SB
899 1513 140 (denotes ) 155 SB
1054 1513 52 (the) 52 SB
165 1563 185 (associated ) 208 SB
373 1563 175 (classifier, ) 198 SB
571 1563 62 (N2 ) 85 SB
656 1563 39 (is ) 62 SB
718 1563 30 (a ) 54 SB
772 1563 140 (number ) 164 SB
936 1563 170 (indicating) 170 SB
165 1613 147 (whether ) 148 SB
313 1613 65 (CL ) 66 SB
379 1613 39 (is ) 40 SB
419 1613 30 (a ) 31 SB
450 1613 77 (unit ) 78 SB
528 1613 46 (or ) 47 SB
575 1613 177 (collective ) 178 SB
753 1613 164 (classifier ) 166 SB
919 1613 46 (\(1 ) 48 SB
967 1613 60 (for ) 62 SB
1029 1613 77 (unit,) 77 SB
165 1663 32 (2 ) 33 SB
198 1663 60 (for ) 61 SB
259 1663 191 (collective\) ) 192 SB
451 1663 72 (and ) 74 SB
525 1663 88 (Freq ) 90 SB
615 1663 140 (denotes ) 142 SB
757 1663 63 (the ) 65 SB
822 1663 180 (frequency ) 182 SB
1004 1663 46 (of ) 48 SB
1052 1663 54 (co-) 54 SB
165 1713 197 (occurrence ) 201 SB
366 1713 152 (between ) 156 SB
522 1713 51 (W ) 55 SB
577 1713 72 (and ) 76 SB
653 1713 76 (CL. ) 80 SB
733 1713 77 (The ) 81 SB
814 1713 162 (semantic ) 166 SB
980 1713 93 (class ) 98 SB
1078 1713 28 (is) 28 SB
165 1763 164 (shown in ) 164 SB
32 0 0 42 42 0 0 0 41 /Times-Bold /font29 ANSIFont font
329 1760 134 (Fig. 3.2) 134 SB
32 0 0 42 42 0 0 0 38 /Times-Roman /font32 ANSIFont font
463 1763 11 (.) 11 SB
32 0 0 42 42 0 0 0 41 /Times-Bold /font29 ANSIFont font
165 1861 111 (Step 5) 111 SB
32 0 0 42 42 0 0 0 38 /Times-Roman /font32 ANSIFont font
276 1864 12 (:) 12 SB
315 1864 592 (Determine representative classifier) 592 SB
165 1915 101 (Input:) 101 SB
315 1915 41 (A ) 65 SB
380 1915 63 (list ) 88 SB
468 1915 46 (of ) 71 SB
539 1915 262 (noun-classifier ) 287 SB
826 1915 86 (with ) 111 SB
937 1915 169 (frequency) 169 SB
165 1965 508 (information of co-occurrence.) 508 SB
165 2015 129 (Output:) 129 SB
315 2015 263 (Representative ) 277 SB
592 2015 164 (classifier ) 178 SB
770 2015 46 (of ) 61 SB
831 2015 89 (each ) 104 SB
935 2015 95 (noun ) 110 SB
1045 2015 61 (and) 61 SB
165 2065 501 (each semantic class of nouns.) 501 SB
315 2165 57 (As ) 61 SB
376 2165 35 (it ) 39 SB
415 2165 70 (can ) 74 SB
489 2165 51 (be ) 56 SB
545 2165 163 (observed ) 168 SB
713 2165 44 (in ) 49 SB
32 0 0 42 42 0 0 0 41 /Times-Bold /font29 ANSIFont font
762 2162 81 (Fig. ) 86 SB
848 2162 53 (3.1) 53 SB
32 0 0 42 42 0 0 0 38 /Times-Roman /font32 ANSIFont font
901 2165 22 (, ) 27 SB
928 2165 89 (each ) 94 SB
1022 2165 84 (noun) 84 SB
165 2216 84 (may ) 102 SB
267 2216 51 (be ) 69 SB
336 2216 88 (used ) 106 SB
442 2216 86 (with ) 104 SB
546 2216 131 (several ) 149 SB
695 2216 149 (possible ) 167 SB
862 2216 191 (classifiers. ) 209 SB
1071 2216 35 (In) 35 SB
165 2266 164 (language ) 177 SB
342 2266 190 (generation ) 203 SB
545 2266 148 (process. ) 161 SB
706 2266 176 (However, ) 189 SB
895 2266 60 (we ) 73 SB
968 2266 91 (have ) 105 SB
1073 2266 33 (to) 33 SB
165 2316 108 (select ) 112 SB
277 2316 86 (only ) 91 SB
368 2316 72 (one ) 77 SB
445 2316 46 (of ) 51 SB
496 2316 107 (them. ) 112 SB
608 2316 69 (For ) 74 SB
682 2316 89 (each ) 94 SB
776 2316 95 (noun ) 100 SB
876 2316 60 (we ) 65 SB
941 2316 108 (select ) 113 SB
1054 2316 52 (the) 52 SB
165 2366 164 (classifier ) 180 SB
345 2366 86 (with ) 102 SB
447 2366 63 (the ) 79 SB
526 2366 143 (greatest ) 159 SB
685 2366 103 (value ) 119 SB
804 2366 46 (of ) 62 SB
866 2366 240 (co-occurrence) 240 SB
165 2416 180 (frequency ) 187 SB
352 2416 44 (to ) 52 SB
404 2416 51 (be ) 59 SB
463 2416 63 (the ) 71 SB
534 2416 249 (representative ) 257 SB
791 2416 164 (classifier ) 172 SB
963 2416 60 (for ) 68 SB
1031 2416 75 (both) 75 SB
165 2466 249 (representative ) 284 SB
449 2466 77 (unit ) 112 SB
561 2466 164 (classifier ) 199 SB
760 2466 72 (and ) 108 SB
868 2466 238 (representative) 238 SB
1210 210 177 (collective ) 191 SB
1401 210 175 (classifier. ) 189 SB
1590 210 77 (The ) 91 SB
1681 210 164 (classifier ) 178 SB
1859 210 44 (in ) 58 SB
32 0 0 42 42 0 0 0 41 /Times-Bold /font29 ANSIFont font
1917 207 81 (Fig. ) 95 SB
2012 207 53 (3.1) 53 SB
32 0 0 42 42 0 0 0 38 /Times-Roman /font32 ANSIFont font
2065 210 22 (, ) 37 SB
2102 210 49 (for) 49 SB
1210 261 166 (example, ) 179 SB
1389 261 77 (will ) 90 SB
1479 261 91 (have ) 104 SB

/MSTT31c15e [42.0 0 0 0 0 0] 23 -23 [-42.0 -42.0 42.0 42.0] [1 42 div 0 0 1 42 div 0 0] /MSTT31c15e GreNewFont

32 0 0 42 42 0 0 1 36 /MSTT31c15e font

/Ga4 [18.0 0.0 1.0 0.0 15.0 17.0]
/Ga4 {
    14 17 true [1 0 0 -1 -1.0 17.0] {<0fc01ff03838601c600cc60ccf0cd98c598c7f0c380c380c180c180c180c180c180c>} imagemask 
  }
  164 /Ga4 MSTT31c15e AddChar
/Gb9 [19.0 0.0 1.0 0.0 19.0 17.0]
/Gb9 {
    18 17 true [1 0 0 -1 -1.0 17.0] {<380c007c0c00cc0c00cc0c007c0c003c0c000c0c000c0c000c0c000c0e000c3f800cfd800dccc00f
8cc00f0c800e0f800c0700>} imagemask 
  }
  185 /Gb9 MSTT31c15e AddChar

1583 263 37 (\244\271) 37 SB
32 0 0 42 42 0 0 0 38 /Times-Roman /font32 ANSIFont font
1620 261 53 (_1 ) 66 SB
1686 261 46 (as ) 59 SB
1745 261 63 (the ) 77 SB
1822 261 249 (representative ) 263 SB
2085 261 66 (unit) 66 SB
1210 318 164 (classifier ) 190 SB
1400 318 72 (and ) 98 SB
1498 318 91 (have ) 117 SB
32 0 0 42 42 0 0 1 36 /MSTT31c15e font

/Gb3 [25.0 0.0 2.0 0.0 25.0 17.0]
/Gb3 {
    23 17 true [1 0 0 -1 -2.0 17.0] {<0fc0603ff060f838607818601c18601e18603818603018603018603018603018fc3019fc3e1b6632
1e66331c64361c7c1e1838>} imagemask 
  }
  179 /Gb3 MSTT31c15e AddChar
/Gd0 [14.0 0.0 2.0 0.0 12.0 16.0]
/Gd0 {
    10 16 true [1 0 0 -1 -2.0 16.0] {<7040c84088c0cb807f00180000000000000000007040d84088c0cb807f003800>} imagemask 
  }
  208 /Gd0 MSTT31c15e AddChar

1615 320 57 (\244\263\320) 57 SB
32 0 0 42 42 0 0 0 38 /Times-Roman /font32 ANSIFont font
1672 318 53 (_2 ) 79 SB
1751 318 46 (as ) 72 SB
1823 318 63 (the ) 90 SB
1913 318 238 (representative) 238 SB
1210 375 177 (collective ) 181 SB
1391 375 72 (one ) 76 SB
1467 375 60 (for ) 64 SB
1531 375 63 (the ) 67 SB
1598 375 95 (noun ) 100 SB
32 0 0 42 42 0 0 1 36 /MSTT31c15e font

/Ga1 [18.0 0.0 1.0 0.0 15.0 17.0]
/Ga1 {
    14 17 true [1 0 0 -1 -1.0 17.0] {<0fe03ff0f838780c1c0c1e0c380c300c300c300c300c300c300c300c300c300c300c>} imagemask 
  }
  161 /Ga1 MSTT31c15e AddChar
/Gc3 [14.0 0.0 0.0 0.0 13.0 17.0]
/Gc3 {
    13 17 true [1 0 0 -1 0.0 17.0] {<0f103ff870787e008f8001c000600060006000600060006003e006600660036003c0>} imagemask 
  }
  195 /Gc3 MSTT31c15e AddChar
/Gc1 [18.0 0.0 2.0 0.0 16.0 17.0]
/Gc1 {
    14 17 true [1 0 0 -1 -2.0 17.0] {<380c7c0ccc0ccc0c7c0c3c0c0c0c0c0c0c0c0c0c3c0c7e0c4f0ccd8c4ccc7c7c3838>} imagemask 
  }
  193 /Gc1 MSTT31c15e AddChar
/Gd2 [14.0 0.0 0.0 0.0 12.0 17.0]
/Gd2 {
    12 17 true [1 0 0 -1 0.0 17.0] {<1f803fe07060c0304030003000300030003000300030003000300030003000300030>} imagemask 
  }
  210 /Gd2 MSTT31c15e AddChar

1698 377 167 (\244\263\320\241\303\303\301\241\322\303) 167 SB
32 0 0 42 42 0 0 0 38 /Times-Roman /font32 ANSIFont font
1865 375 106 (_111. ) 111 SB
1976 375 175 (Collective) 175 SB
1210 432 180 (classifiers ) 182 SB
1392 432 63 (are ) 65 SB
1457 432 88 (used ) 91 SB
1548 432 131 (instead ) 134 SB
1682 432 46 (of ) 49 SB
1731 432 77 (unit ) 80 SB
1811 432 180 (classifiers ) 183 SB
1994 432 102 (when ) 105 SB
2099 432 52 (the) 52 SB
1210 482 487 (notion of `group' is required.) 487 SB
1360 532 70 (We ) 76 SB
1436 532 79 (also ) 85 SB
1521 532 79 (find ) 85 SB
1606 532 63 (the ) 69 SB
1675 532 249 (representative ) 256 SB
1931 532 164 (classifier ) 171 SB
2102 532 49 (for) 49 SB
1210 582 89 (each ) 90 SB
1300 582 162 (semantic ) 163 SB
1463 582 93 (class ) 94 SB
1557 582 46 (of ) 48 SB
1605 582 111 (nouns ) 113 SB
1718 582 44 (in ) 46 SB
1764 582 63 (the ) 65 SB
1829 582 98 (same ) 100 SB
1929 582 149 (manner. ) 151 SB
2080 582 11 ( ) 13 SB
2093 582 58 (For) 58 SB
1210 632 89 (each ) 90 SB
1300 632 162 (semantic ) 163 SB
1463 632 93 (class ) 94 SB
1557 632 46 (of ) 47 SB
1604 632 111 (nouns ) 112 SB
1716 632 163 (\(grouped ) 164 SB
1880 632 53 (by ) 55 SB
1935 632 63 (the ) 65 SB
2000 632 151 (semantic) 151 SB
1210 682 93 (class ) 97 SB
1307 682 153 (attached ) 157 SB
1464 682 86 (with ) 90 SB
1554 682 89 (each ) 93 SB
1647 682 120 (noun\), ) 124 SB
1771 682 63 (the ) 68 SB
1839 682 164 (classifier ) 169 SB
2008 682 86 (with ) 91 SB
2099 682 52 (the) 52 SB
1210 732 143 (greatest ) 150 SB
1360 732 103 (value ) 110 SB
1470 732 46 (of ) 53 SB
1523 732 251 (co-occurrence ) 258 SB
1781 732 180 (frequency ) 187 SB
1968 732 39 (is ) 46 SB
2014 732 137 (selected) 137 SB
1210 782 44 (to ) 59 SB
1269 782 51 (be ) 66 SB
1335 782 63 (the ) 78 SB
1413 782 260 (representative. ) 275 SB
1688 782 77 (The ) 92 SB
1780 782 164 (classifier ) 179 SB
1959 782 39 (is ) 55 SB
2014 782 88 (used ) 104 SB
2118 782 33 (to) 33 SB
1210 832 124 (handle ) 125 SB
1335 832 63 (the ) 64 SB
1399 832 201 (assignment ) 202 SB
1601 832 46 (of ) 48 SB
1649 832 164 (classifier ) 166 SB
1815 832 44 (to ) 46 SB
1861 832 95 (noun ) 97 SB
1958 832 114 (which ) 116 SB
2074 832 77 (does) 77 SB
1210 882 65 (not ) 81 SB
1291 882 91 (exist ) 107 SB
1398 882 44 (in ) 60 SB
1458 882 63 (the ) 79 SB
1537 882 129 (trained ) 145 SB
1682 882 134 (corpus. ) 150 SB
1832 882 69 (For ) 85 SB
1917 882 166 (example, ) 182 SB
2099 882 52 (the) 52 SB
1210 932 249 (representative ) 256 SB
1466 932 77 (unit ) 84 SB
1550 932 180 (classifiers ) 187 SB
1737 932 60 (for ) 67 SB
1804 932 89 (each ) 96 SB
1900 932 162 (semantic ) 169 SB
2069 932 82 (class) 82 SB
1210 982 167 (extracted ) 168 SB
1378 982 53 (by ) 54 SB
1432 982 63 (the ) 64 SB
1496 982 129 (pattern ) 130 SB
1626 982 69 (\(N- ) 71 SB
1697 982 232 (-NCNM-CL\) ) 234 SB
1931 982 63 (are ) 65 SB
1996 982 120 (shown ) 122 SB
2118 982 33 (in) 33 SB
32 0 0 42 42 0 0 0 41 /Times-Bold /font29 ANSIFont font
1210 1029 134 (Fig. 3.3) 134 SB
32 0 0 42 42 0 0 0 38 /Times-Roman /font32 ANSIFont font
1344 1032 11 (.) 11 SB
32 0 0 46 46 0 0 0 44 /Times-Bold /font29 ANSIFont font
1210 1230 455 (4. Classifier Resolution) 455 SB
32 0 0 42 42 0 0 0 38 /Times-Roman /font32 ANSIFont font
1210 1339 77 (The ) 84 SB
1294 1339 215 (associations ) 222 SB
1516 1339 46 (as ) 53 SB
1569 1339 168 (produced ) 175 SB
1744 1339 44 (in ) 52 SB
1796 1339 63 (the ) 71 SB
1867 1339 156 (previous ) 164 SB
2031 1339 120 (section) 120 SB
1210 1389 63 (are ) 74 SB
1284 1389 114 (useful ) 125 SB
1409 1389 60 (for ) 72 SB
1481 1389 216 (determining ) 228 SB
1709 1389 30 (a ) 42 SB
1751 1389 121 (proper ) 133 SB
1884 1389 164 (classifier ) 176 SB
2060 1389 60 (for ) 72 SB
2132 1389 19 (a) 19 SB
1210 1439 105 (given ) 118 SB
1328 1439 106 (noun. ) 119 SB
1447 1439 11 ( ) 24 SB
1471 1439 69 (For ) 82 SB
1553 1439 30 (a ) 44 SB
1597 1439 95 (noun ) 109 SB
1706 1439 173 (occurring ) 187 SB
1893 1439 44 (in ) 58 SB
1951 1439 63 (the ) 77 SB
2028 1439 123 (corpus,) 123 SB
1210 1489 191 (alternative ) 222 SB
1432 1489 247 (determination ) 278 SB
1710 1489 39 (is ) 70 SB
1780 1489 244 (accomplished ) 276 SB
2056 1489 44 (in ) 76 SB
2132 1489 19 (a) 19 SB
1210 1539 271 (straightforward ) 301 SB
1511 1539 138 (manner ) 168 SB
1679 1539 53 (by ) 83 SB
1762 1539 102 (using ) 133 SB
1895 1539 51 (its ) 82 SB
1977 1539 174 (associated) 174 SB
1210 1589 249 (representative ) 261 SB
1471 1589 164 (classifier ) 176 SB
1647 1589 114 (which ) 126 SB
1773 1589 121 (occurs ) 133 SB
1906 1589 44 (in ) 57 SB
1963 1589 63 (the ) 76 SB
2039 1589 112 (corpus) 112 SB
1210 1639 98 (more ) 100 SB
1310 1639 185 (frequently ) 187 SB
1497 1639 84 (than ) 86 SB
1583 1639 72 (any ) 74 SB
1657 1639 98 (other ) 100 SB
1757 1639 191 (classifiers. ) 193 SB
1950 1639 46 (In ) 48 SB
1998 1639 63 (the ) 66 SB
2064 1639 87 (other) 87 SB
1210 1689 461 (case where the given noun ) 462 SB
1672 1689 88 (does ) 89 SB
1761 1689 65 (not ) 66 SB
1827 1689 91 (exist ) 92 SB
1919 1689 44 (in ) 45 SB
1964 1689 63 (the ) 64 SB
2028 1689 123 (corpus,) 123 SB
1210 1739 63 (the ) 69 SB
1279 1739 247 (determination ) 253 SB
1532 1739 39 (is ) 45 SB
1577 1739 93 (done ) 99 SB
1676 1739 53 (by ) 59 SB
1735 1739 102 (using ) 108 SB
1843 1739 63 (the ) 70 SB
1913 1739 238 (representative) 238 SB
1210 1789 775 (classifier of its class in the concept hierarchy.) 775 SB
1360 1889 107 (Some ) 114 SB
1474 1889 171 (examples ) 178 SB
1652 1889 46 (of ) 53 SB
1705 1889 164 (classifier ) 171 SB
1876 1889 216 (determining ) 223 SB
2099 1889 52 (are) 52 SB
1210 1939 103 (listed ) 114 SB
1324 1939 125 (below. ) 136 SB
1460 1939 11 ( ) 23 SB
1483 1939 60 (\(1\) ) 72 SB
1555 1939 72 (and ) 84 SB
1639 1939 60 (\(3\) ) 72 SB
1711 1939 99 (show ) 111 SB
1822 1939 63 (the ) 75 SB
1897 1939 84 (case ) 96 SB
1993 1939 46 (of ) 58 SB
2051 1939 100 (nouns) 100 SB
1210 1989 178 (appearing ) 189 SB
1399 1989 44 (in ) 56 SB
1455 1989 63 (the ) 75 SB
1530 1989 134 (corpus, ) 146 SB
1676 1989 105 (while ) 117 SB
1793 1989 60 (\(2\) ) 72 SB
1865 1989 72 (and ) 84 SB
1949 1989 60 (\(4\) ) 72 SB
2021 1989 99 (show ) 111 SB
2132 1989 19 (a) 19 SB
1210 2039 157 (different ) 158 SB
1368 2039 163 (scenario. ) 165 SB
1533 2039 46 (In ) 48 SB
1581 2039 71 (\(2\), ) 73 SB
1654 2039 63 (the ) 65 SB
1719 2039 77 (unit ) 79 SB
1798 2039 164 (classifier ) 166 SB
1964 2039 46 (of ) 48 SB
2012 2039 139 (/appern/) 139 SB
1210 2089 740 (is obtained by using the representative unit ) 741 SB
1951 2089 164 (classifier ) 165 SB
2116 2089 35 (of) 35 SB
1210 2141 51 (its ) 56 SB
1266 2141 93 (class ) 98 SB
1364 2141 106 (`fruit' ) 111 SB
1475 2141 114 (which ) 119 SB
1594 2141 39 (is ) 44 SB
32 0 0 50 50 0 0 1 43 /MSTT31c12c font
1638 2136 30 (\305\331 ) 19 SB
1657 2136 61 (\241_1) 61 SB
32 0 0 42 42 0 0 0 38 /Times-Roman /font32 ANSIFont font
1718 2141 11 ( ) 16 SB
1734 2141 110 (/luuk/ ) 115 SB
1849 2141 178 (according ) 183 SB
2032 2141 44 (to ) 49 SB
32 0 0 42 42 0 0 0 41 /Times-Bold /font29 ANSIFont font
2081 2138 70 (Fig.) 70 SB
1210 2198 53 (3.3) 53 SB
32 0 0 42 42 0 0 0 38 /Times-Roman /font32 ANSIFont font
1263 2201 22 (. ) 35 SB
1298 2201 180 (Similarly, ) 193 SB
1491 2201 44 (in ) 57 SB
1548 2201 71 (\(4\), ) 85 SB
1633 2201 63 (the ) 77 SB
1710 2201 177 (collective ) 191 SB
1901 2201 164 (classifier ) 178 SB
2079 2201 46 (of ) 60 SB
2139 2201 12 (/) 12 SB
1210 2252 460 (gangken/ is determined by ) 461 SB
1671 2252 63 (the ) 64 SB
1735 2252 249 (representative ) 250 SB
1985 2252 166 (collective) 166 SB
1210 2304 656 (classifier of its class `animal' which is ) 656 SB
32 0 0 50 50 0 0 1 43 /MSTT31c12c font
1866 2299 32 (\275\331 ) 21 SB
1887 2299 56 (\247_2) 56 SB
32 0 0 42 42 0 0 0 38 /Times-Roman /font32 ANSIFont font
1943 2304 144 ( /fuung/.) 144 SB
0 0 0 fC
/fm 256 def
2 2 387 2663 B
1 F
n
/fm 256 def
2 2 387 2663 B
1 F
n
/fm 256 def
446 2 390 2663 B
1 F
n
/fm 256 def
2 2 837 2663 B
1 F
n
/fm 256 def
446 2 840 2663 B
1 F
n
/fm 256 def
2 2 1287 2663 B
1 F
n
/fm 256 def
446 2 1290 2663 B
1 F
n
/fm 256 def
2 2 1737 2663 B
1 F
n
/fm 256 def
2 2 1737 2663 B
1 F
n
/fm 256 def
2 49 387 2666 B
1 F
n
/fm 256 def
2 49 837 2666 B
1 F
n
/fm 256 def
2 49 1287 2666 B
1 F
n
/fm 256 def
2 49 1737 2666 B
1 F
n
gs 447 53 390 2663 CB
482 2669 251 (Semantic class) 251 SB
gr
gs 447 53 840 2663 CB
938 2669 239 (Unit classifier) 239 SB
gr
gs 447 53 1290 2663 CB
1338 2669 339 (Collective classifier) 339 SB
gr
/fm 256 def
2 2 387 2716 B
1 F
n
/fm 256 def
446 2 390 2716 B
1 F
n
/fm 256 def
2 2 837 2716 B
1 F
n
/fm 256 def
446 2 840 2716 B
1 F
n
/fm 256 def
2 2 1287 2716 B
1 F
n
/fm 256 def
446 2 1290 2716 B
1 F
n
/fm 256 def
2 2 1737 2716 B
1 F
n
/fm 256 def
2 61 387 2719 B
1 F
n
/fm 256 def
2 61 837 2719 B
1 F
n
/fm 256 def
2 61 1287 2719 B
1 F
n
/fm 256 def
2 61 1737 2719 B
1 F
n
gs 447 53 390 2716 CB
549 2722 116 (animal) 116 SB
gr
32 0 0 50 50 0 0 1 43 /MSTT31c12c font
gs 447 65 840 2716 CB
1018 2719 32 (\265\321 ) 21 SB
1039 2719 58 (\307_1) 58 SB
gr
gs 447 65 1290 2716 CB
1469 2719 32 (\275\331 ) 21 SB
1490 2719 56 (\247_2) 56 SB
gr
/fm 256 def
2 61 387 2781 B
1 F
n
/fm 256 def
2 61 837 2781 B
1 F
n
/fm 256 def
2 61 1287 2781 B
1 F
n
/fm 256 def
2 61 1737 2781 B
1 F
n
32 0 0 42 42 0 0 0 38 /Times-Roman /font32 ANSIFont font
gs 447 50 390 2781 CB
550 2784 115 (human) 115 SB
gr
32 0 0 50 50 0 0 1 43 /MSTT31c12c font
gs 447 62 840 2781 CB
1016 2781 83 (\244\271_1) 83 SB
gr
gs 447 62 1290 2781 CB
1454 2781 107 (\244\263\320_2) 107 SB
gr
/fm 256 def
2 61 387 2843 B
1 F
n
/fm 256 def
2 61 837 2843 B
1 F
n
/fm 256 def
2 61 1287 2843 B
1 F
n
/fm 256 def
2 61 1737 2843 B
1 F
n
32 0 0 42 42 0 0 0 38 /Times-Roman /font32 ANSIFont font
gs 447 50 390 2843 CB
565 2846 85 (plant) 85 SB
gr
32 0 0 50 50 0 0 1 43 /MSTT31c12c font
gs 447 62 840 2843 CB
1016 2843 32 (\265\214 ) 21 SB
1037 2843 62 (\271_1) 62 SB
gr

/G2d [20.0 0.0 4.0 9.0 16.0 11.0]
/G2d {
    12 2 true [1 0 0 -1 -4.0 11.0] {<fff0fff0>} imagemask 
  }
  45 /G2d MSTT31c12c AddChar

gs 447 62 1290 2843 CB
1497 2843 20 (-) 20 SB
gr
/fm 256 def
2 61 387 2905 B
1 F
n
/fm 256 def
2 2 387 2967 B
1 F
n
/fm 256 def
2 2 387 2967 B
1 F
n
/fm 256 def
446 2 390 2967 B
1 F
n
/fm 256 def
2 61 837 2905 B
1 F
n
/fm 256 def
2 2 837 2967 B
1 F
n
/fm 256 def
446 2 840 2967 B
1 F
n
/fm 256 def
2 61 1287 2905 B
1 F
n
/fm 256 def
2 2 1287 2967 B
1 F
n
/fm 256 def
446 2 1290 2967 B
1 F
n
/fm 256 def
2 61 1737 2905 B
1 F
n
/fm 256 def
2 2 1737 2967 B
1 F
n
/fm 256 def
2 2 1737 2967 B
1 F
n
32 0 0 42 42 0 0 0 38 /Times-Roman /font32 ANSIFont font
gs 447 50 390 2905 CB
571 2908 73 (fruit) 73 SB
gr
32 0 0 50 50 0 0 1 43 /MSTT31c12c font
gs 447 62 840 2905 CB
1017 2905 30 (\305\331 ) 19 SB
1036 2905 61 (\241_1) 61 SB
gr
gs 447 62 1290 2905 CB
1497 2905 20 (-) 20 SB
gr
32 0 0 42 42 0 0 0 41 /Times-Bold /font29 ANSIFont font
805 3020 134 (Fig. 3.3) 134 SB
32 0 0 42 42 0 0 0 38 /Times-Roman /font32 ANSIFont font
939 3023 572 ( NCA for representative classifier) 572 SB
1 #C
statusdict begin /manualfeed false store end
EJ RS
%%PageTrailer
%%PageResources: font MSTT31c12c
%%+ font MSTT31c15e
%%+ font Times-Bold
%%+ font Times-Italic
%%+ font Times-Roman
%%Page: 6 6
%%PageResources: (atend)
SS
0 0 25 31 776 1169 300 SM
32 0 0 42 42 0 0 0 41 /Times-Bold /font29 ANSIFont font
0 0 0 fC
165 407 305 (     Unit classifier) 305 SB
32 0 0 42 42 0 0 0 38 /Times-Roman /font32 ANSIFont font
315 467 71 (\(1\)  ) 71 SB
32 0 0 50 50 0 0 1 47 /MSTT31c146 font
386 458 34 (\271\321 ) 23 SB
409 458 56 (\241\340\303\325 ) 45 SB
454 458 226 (\302\271     \244\271     \267\325\350 ) 215 SB
669 458 88 (     \312\325\350 ) 77 SB
32 0 0 42 42 0 0 0 38 /Times-Roman /font32 ANSIFont font
315 528 499 (      /nakrian    kon     tii    sii/) 499 SB
315 578 612 (       student <student> number  four) 612 SB
315 634 71 (\(2\)  ) 71 SB
32 0 0 50 50 0 0 1 47 /MSTT31c146 font

/G81 [0.0 0.0 -23.0 24.0 -7.0 32.0]
/G81 {
    16 8 true [1 0 0 -1 23.0 32.0] {<07e01ff8381c60064002ffe3003f0003>} imagemask 
  }
  129 /G81 MSTT31c146 AddChar
/G9c [0.0 0.0 -12.0 37.0 -1.0 47.0]
/G9c {
    11 10 true [1 0 0 -1 12.0 47.0] {<006070604c608c404cc03cc009801b007e00f000>} imagemask 
  }
  156 /G9c MSTT31c146 AddChar
/Ge4 [17.0 0.0 2.0 0.0 15.0 35.0]
/Ge4 {
    13 35 true [1 0 0 -1 -2.0 35.0] {<8000600063c066e064e064e064e064e064e064e064e038e000e000e000e000e000e000e000e000e0
00e000e000e000e000e000e000e000e000e000f001880108010800880070>} imagemask 
  }
  228 /Ge4 MSTT31c146 AddChar

386 625 102 (\341\315\273\340\273\201\234 ) 91 SB
477 625 106 (\305     \305\331 ) 95 SB
572 625 138 (\241     \344\313\271) 138 SB
32 0 0 42 42 0 0 0 38 /Times-Roman /font32 ANSIFont font
315 695 420 (      /appern   luuk     nai/) 420 SB
315 745 467 (       apple   <apple>  which) 467 SB
32 0 0 42 42 0 0 0 41 /Times-Bold /font29 ANSIFont font
165 843 394 (    Collective classifier) 394 SB
32 0 0 42 42 0 0 0 38 /Times-Roman /font32 ANSIFont font
315 903 60 (\(3\) ) 60 SB
32 0 0 50 50 0 0 1 47 /MSTT31c146 font

/Gb3 [27.0 0.0 2.0 0.0 24.0 20.0]
/Gb3 {
    22 20 true [1 0 0 -1 -2.0 20.0] {<1f800838e038607038c03838803838fe383838383838383838383838383838383838383838383838
38383f38f0393b7c38be4438bcc419384c0e3878>} imagemask 
  }
  179 /Gb3 MSTT31c146 AddChar

375 894 444 ( \244\263\320\241\303\303\301\241\322\303         \244\263\320    \271\321\351 ) 433 SB
808 894 23 (\271) 23 SB
32 0 0 42 42 0 0 0 38 /Times-Roman /font32 ANSIFont font
315 964 566 (      /kanagammagarn  kana   nan/) 566 SB
315 1014 573 (        committee           group  that) 573 SB
315 1070 71 (\(4\)  ) 71 SB
32 0 0 50 50 0 0 1 47 /MSTT31c146 font

/Ga2 [20.0 0.0 1.0 0.0 17.0 20.0]
/Ga2 {
    16 20 true [1 0 0 -1 -1.0 20.0] {<0e013f8363c741c779c7edc7c5c745c739c701c701c701c701c701c701c701c701c701c700ee007c
>} imagemask 
  }
  162 /Ga2 MSTT31c146 AddChar

386 1061 208 ( \241\322\247\340\242\271      \275\331 ) 197 SB
583 1061 116 (\247      \271\321\351 ) 105 SB
688 1061 23 (\271) 23 SB
32 0 0 42 42 0 0 0 38 /Times-Roman /font32 ANSIFont font
315 1131 447 (       /gangken  fuung  nan/) 447 SB
315 1181 441 (        magpie    group  that) 441 SB
32 0 0 46 46 0 0 0 44 /Times-Bold /font29 ANSIFont font
165 1328 268 (5. Conclusion) 268 SB
32 0 0 42 42 0 0 0 38 /Times-Roman /font32 ANSIFont font
165 1437 477 (The proposed approach is a ) 478 SB
643 1437 223 (significantly ) 224 SB
867 1437 81 (new ) 82 SB
949 1437 138 (method ) 139 SB
1088 1437 33 (to) 33 SB
165 1487 200 (manipulate ) 204 SB
369 1487 63 (the ) 68 SB
437 1487 164 (classifier ) 169 SB
606 1487 121 (phrase ) 126 SB
732 1487 44 (in ) 49 SB
781 1487 89 (Thai ) 94 SB
875 1487 175 (language. ) 180 SB
1055 1487 66 (The) 66 SB
165 1537 75 (fact ) 81 SB
246 1537 75 (that ) 81 SB
327 1537 63 (the ) 69 SB
396 1537 191 (expression ) 197 SB
593 1537 46 (of ) 52 SB
645 1537 100 (some ) 106 SB
751 1537 162 (syntactic ) 168 SB
919 1537 202 (constituents) 202 SB
165 1587 903 (needs a specific classifier to be constructed with and ) 904 SB
1069 1587 52 (the) 52 SB
165 1637 162 (selection ) 173 SB
338 1637 46 (of ) 57 SB
395 1637 164 (classifier ) 175 SB
570 1637 60 (for ) 71 SB
641 1637 89 (each ) 100 SB
741 1637 95 (noun ) 106 SB
847 1637 46 (or ) 57 SB
904 1637 95 (noun ) 107 SB
1011 1637 110 (phrase) 110 SB
165 1687 149 (depends ) 155 SB
320 1687 53 (on ) 59 SB
379 1687 63 (the ) 69 SB
448 1687 186 (traditional ) 192 SB
640 1687 67 (use ) 73 SB
713 1687 72 (and ) 78 SB
791 1687 63 (the ) 69 SB
860 1687 162 (semantic ) 168 SB
1028 1687 93 (class.) 93 SB
165 1737 77 (The ) 102 SB
267 1737 233 (corpus-based ) 258 SB
525 1737 166 (approach ) 191 SB
716 1737 39 (is ) 65 SB
781 1737 96 (quite ) 122 SB
903 1737 143 (suitable ) 169 SB
1072 1737 49 (for) 49 SB
165 1787 167 (detecting ) 169 SB
334 1787 63 (the ) 66 SB
400 1787 186 (traditional ) 189 SB
589 1787 67 (use ) 70 SB
659 1787 72 (and ) 75 SB
734 1787 173 (searching ) 176 SB
910 1787 60 (for ) 63 SB
973 1787 63 (the ) 66 SB
1039 1787 82 (most) 82 SB
165 1837 501 (appropriate one when it does ) 502 SB
667 1837 65 (not ) 66 SB
733 1837 91 (exist ) 92 SB
825 1837 44 (in ) 45 SB
870 1837 63 (the ) 64 SB
934 1837 123 (corpus ) 124 SB
1058 1837 63 (yet.) 63 SB
165 1887 152 (Concept ) 161 SB
326 1887 171 (hierarchy ) 180 SB
506 1887 46 (of ) 55 SB
561 1887 95 (noun ) 104 SB
665 1887 156 (provides ) 165 SB
830 1887 138 (another ) 148 SB
978 1887 84 (path ) 94 SB
1072 1887 49 (for) 49 SB
165 1937 173 (searching ) 180 SB
345 1937 102 (when ) 109 SB
454 1937 63 (the ) 71 SB
525 1937 99 (NCA ) 107 SB
632 1937 88 (does ) 96 SB
728 1937 65 (not ) 73 SB
801 1937 105 (cover ) 113 SB
914 1937 63 (the ) 71 SB
985 1937 95 (noun ) 103 SB
1088 1937 33 (in) 33 SB
165 1987 154 (question.) 154 SB
315 2037 46 (In ) 48 SB
363 2037 63 (the ) 65 SB
428 2037 123 (future, ) 125 SB
553 2037 72 (this ) 74 SB
627 2037 99 (NCA ) 101 SB
728 2037 77 (will ) 80 SB
808 2037 51 (be ) 54 SB
862 2037 157 (included ) 160 SB
1022 2037 44 (in ) 47 SB
1069 2037 52 (the) 52 SB
165 2087 190 (generation ) 193 SB
358 2087 137 (process ) 140 SB
498 2087 46 (of ) 49 SB
547 2087 159 (Machine ) 162 SB
709 2087 204 (Translation ) 208 SB
917 2087 44 (to ) 48 SB
965 2087 100 (solve ) 104 SB
1069 2087 52 (the) 52 SB
165 2137 164 (classifier ) 170 SB
335 2137 212 (assignment, ) 218 SB
553 2137 72 (and ) 79 SB
632 2137 225 (incorporated ) 232 SB
864 2137 44 (in ) 51 SB
915 2137 63 (the ) 70 SB
985 2137 136 (analysis) 136 SB
165 2187 137 (process ) 150 SB
315 2187 44 (to ) 57 SB
372 2187 147 (produce ) 160 SB
532 2187 30 (a ) 43 SB
575 2187 121 (proper ) 134 SB
709 2187 162 (syntactic ) 175 SB
884 2187 72 (and ) 86 SB
970 2187 151 (semantic) 151 SB
165 2237 170 (structure. ) 175 SB
340 2237 77 (The ) 83 SB
423 2237 164 (classifier ) 170 SB
593 2237 77 (will ) 83 SB
676 2237 84 (then ) 90 SB
766 2237 51 (be ) 57 SB
823 2237 30 (a ) 36 SB
859 2237 72 (key ) 78 SB
937 2237 60 (for ) 66 SB
1003 2237 118 (pattern) 118 SB
165 2287 272 (disambiguation ) 276 SB
441 2287 11 ( ) 15 SB
456 2287 102 (when ) 106 SB
562 2287 35 (it ) 39 SB
601 2287 39 (is ) 43 SB
644 2287 98 (fixed ) 102 SB
746 2287 44 (to ) 48 SB
794 2287 72 (one ) 76 SB
870 2287 46 (of ) 50 SB
920 2287 63 (the ) 67 SB
987 2287 134 (patterns) 134 SB
165 2337 225 (illustrated in ) 225 SB
32 0 0 42 42 0 0 0 41 /Times-Bold /font29 ANSIFont font
390 2334 134 (Fig. 2.1) 134 SB
32 0 0 42 42 0 0 0 38 /Times-Roman /font32 ANSIFont font
524 2337 11 (.) 11 SB
32 0 0 46 46 0 0 0 44 /Times-Bold /font29 ANSIFont font
165 2435 362 (Acknowledgement) 362 SB
32 0 0 42 42 0 0 0 38 /Times-Roman /font32 ANSIFont font
165 2544 70 (We ) 93 SB
258 2544 90 (wish ) 113 SB
371 2544 44 (to ) 67 SB
438 2544 105 (thank ) 128 SB
566 2544 63 (the ) 87 SB
653 2544 157 (National ) 181 SB
834 2544 202 (Electronics ) 226 SB
1060 2544 61 (and) 61 SB
165 2594 393 (Computer Technology ) 394 SB
559 2594 124 (Center ) 125 SB
684 2594 203 (\(NECTEC\) ) 204 SB
888 2594 72 (and ) 73 SB
961 2594 124 (Center ) 125 SB
1086 2594 35 (of) 35 SB
165 2644 63 (the ) 88 SB
253 2644 228 (International ) 253 SB
506 2644 220 (Cooperation ) 245 SB
751 2644 60 (for ) 85 SB
836 2644 285 (Computerization) 285 SB
165 2694 137 (\(CICC\) ) 142 SB
307 2694 83 (who ) 88 SB
395 2694 140 (provide ) 145 SB
540 2694 158 (facilities ) 163 SB
703 2694 72 (and ) 77 SB
780 2694 30 (a ) 35 SB
815 2694 96 (large ) 102 SB
917 2694 123 (corpus ) 129 SB
1046 2694 75 (base) 75 SB
165 2744 275 (for the research.) 275 SB
32 0 0 46 46 0 0 0 44 /Times-Bold /font29 ANSIFont font
165 2841 212 (References) 212 SB
32 0 0 42 42 0 0 0 38 /Times-Roman /font32 ANSIFont font
165 2950 896 ([1] Biber, Douglas. \(1993\).  "Co-occurrence Patterns) 896 SB
165 3000 859 (      among Collocations: A Tool for Corpus-Based) 859 SB
165 3050 909 (      Lexical Knowledge Acquisition".  Computational) 909 SB
1195 310 817 (      Linguistics, Vol. 19, No.3, September 1993.) 817 SB
1195 410 858 ([2] Nagao, Makato. \(1993\). "Machine Translation:) 858 SB
1195 460 944 (      What Have We to Do". Proceedings of MT Summit) 944 SB
1195 510 666 (      IV, June 20-22, 1993, Kobe, Japan.) 666 SB
1195 610 939 ([3] Noss, Richard B. \(1964\). Thai Reference Grammar,) 939 SB
1195 660 946 (      U.S. Government Printing Office, Washington, DC.) 946 SB
1195 710 11 (.) 11 SB
1195 760 884 ([4] Smadja, Frank. \(1993\). "Retrieving Collocations) 884 SB
1195 810 878 (      from Text: Xtract". Computational  Linguistics,) 878 SB
1195 860 536 (      Vol. 19, No.1, March 1993.) 536 SB
1195 960 926 ([5] Sornlertlamvanich, Virach. \(1993\), "Word Segmen) 926 SB
2121 960 14 (-) 14 SB
1195 1010 882 (      tation for Thai in Machine Translation System",) 882 SB
1195 1060 122 (      Ma) 122 SB
1317 1060 760 (chine Translation, National   Electronics and) 760 SB
1195 1110 755 (      Computer Technology Center, \(in Thai\).) 755 SB
1 #C
statusdict begin /manualfeed false store end
EJ RS
%%PageTrailer
%%PageResources: font MSTT31c146
%%+ font Times-Bold
%%+ font Times-Roman
%%Trailer
SVDoc restore
end
%%Pages: 6
% TrueType font name key:
%    MSTT31c12c = 2ad7DBrowalliaUPCF00000032000001900000
%    MSTT31c146 = 2ad7DAngsanaUPCF00000032000001900000
%    MSTT31c152 = 2ad7DBrowalliaUPCF00000026000001900000
%    MSTT31c15e = 2ad7DBrowalliaUPCF0000002a000001900000
%%DocumentSuppliedResources: procset Win35Dict 3 1
%%+ font MSTT31c12c
%%+ font MSTT31c146
%%+ font MSTT31c15e

%%DocumentNeededResources: font AvantGarde-Book
%%+ font Times-Bold
%%+ font Times-Italic
%%+ font Times-Roman

%%EOF
